\documentclass{article}

\usepackage[dblblindworkshop,final]{neurips_2026}
\workshoptitle{SLM-Agents: 1st Workshop on Small Language Models for Agentic Systems, Paris, France, 2026}

\usepackage{pgfplots}
\pgfplotsset{compat=1.18}
\usepackage{subcaption}
\usepackage{enumitem}
\usepackage[utf8]{inputenc}
\usepackage[T1]{fontenc}
\usepackage{hyperref}
\usepackage{url}
\usepackage{booktabs}
\usepackage{amsmath}
\usepackage{graphicx}
\usepackage{amssymb}

\makeatletter
\newcommand{\vfc@missing}[2]{\textbf{??#1/#2??}}
\newcommand{\vfcnum}[2]{%
  \ifcsname vfc@#1@#2\endcsname\csname vfc@#1@#2\endcsname\else\vfc@missing{#1}{#2}\fi}
\newcommand{\vfclat}[2]{%
  \ifcsname vfclat@#1@#2\endcsname\csname vfclat@#1@#2\endcsname\else\vfc@missing{#1}{#2}\fi}

\expandafter\gdef\csname vfc@qwen06b-ft@seen\endcsname{0.979 $\pm$ 0.001}
\expandafter\gdef\csname vfc@qwen06b-ft@unseen\endcsname{0.000 $\pm$ 0.000}
\expandafter\gdef\csname vfc@qwen06b-ft@out_of_scope\endcsname{0.831 $\pm$ 0.004}
\expandafter\gdef\csname vfc@qwen06b-ft@seeds\endcsname{3}
\expandafter\gdef\csname vfc@qwen06b-sip@seen\endcsname{0.987 $\pm$ 0.003}
\expandafter\gdef\csname vfc@qwen06b-sip@unseen\endcsname{0.593 $\pm$ 0.043}
\expandafter\gdef\csname vfc@qwen06b-sip@out_of_scope\endcsname{0.984 $\pm$ 0.001}
\expandafter\gdef\csname vfc@qwen06b-sip@seeds\endcsname{3}
\expandafter\gdef\csname vfc@qwen17b-ft@seen\endcsname{0.955 $\pm$ 0.009}
\expandafter\gdef\csname vfc@qwen17b-ft@unseen\endcsname{0.000 $\pm$ 0.000}
\expandafter\gdef\csname vfc@qwen17b-ft@out_of_scope\endcsname{0.826 $\pm$ 0.004}
\expandafter\gdef\csname vfc@qwen17b-ft@seeds\endcsname{3}
\expandafter\gdef\csname vfc@qwen17b-sip@seen\endcsname{0.981 $\pm$ 0.004}
\expandafter\gdef\csname vfc@qwen17b-sip@unseen\endcsname{0.841 $\pm$ 0.016}
\expandafter\gdef\csname vfc@qwen17b-sip@out_of_scope\endcsname{0.957 $\pm$ 0.003}
\expandafter\gdef\csname vfc@qwen17b-sip@seeds\endcsname{3}
\expandafter\gdef\csname vfc@qwen17b-sip-fullft@seen\endcsname{0.990 $\pm$ 0.003}
\expandafter\gdef\csname vfc@qwen17b-sip-fullft@unseen\endcsname{0.754 $\pm$ 0.048}
\expandafter\gdef\csname vfc@qwen17b-sip-fullft@out_of_scope\endcsname{0.986 $\pm$ 0.004}
\expandafter\gdef\csname vfc@qwen17b-sip-fullft@seeds\endcsname{3}
\expandafter\gdef\csname vfc@fg270m-ft@seen\endcsname{0.971 $\pm$ 0.005}
\expandafter\gdef\csname vfc@fg270m-ft@unseen\endcsname{0.000 $\pm$ 0.000}
\expandafter\gdef\csname vfc@fg270m-ft@out_of_scope\endcsname{0.814 $\pm$ 0.015}
\expandafter\gdef\csname vfc@fg270m-ft@seeds\endcsname{3}
\expandafter\gdef\csname vfc@fg270m-sip@seen\endcsname{0.972 $\pm$ 0.004}
\expandafter\gdef\csname vfc@fg270m-sip@unseen\endcsname{0.186 $\pm$ 0.015}
\expandafter\gdef\csname vfc@fg270m-sip@out_of_scope\endcsname{0.973 $\pm$ 0.006}
\expandafter\gdef\csname vfc@fg270m-sip@seeds\endcsname{3}
\expandafter\gdef\csname vfc@fg270m-sip-native@seen\endcsname{0.974 $\pm$ 0.003}
\expandafter\gdef\csname vfc@fg270m-sip-native@unseen\endcsname{0.249 $\pm$ 0.025}
\expandafter\gdef\csname vfc@fg270m-sip-native@out_of_scope\endcsname{0.948 $\pm$ 0.005}
\expandafter\gdef\csname vfc@fg270m-sip-native@seeds\endcsname{3}
\expandafter\gdef\csname vfc@gemma270m-ft@seen\endcsname{0.975 $\pm$ 0.001}
\expandafter\gdef\csname vfc@gemma270m-ft@unseen\endcsname{0.000 $\pm$ 0.000}
\expandafter\gdef\csname vfc@gemma270m-ft@out_of_scope\endcsname{0.816 $\pm$ 0.011}
\expandafter\gdef\csname vfc@gemma270m-ft@seeds\endcsname{3}
\expandafter\gdef\csname vfc@gemma270m-sip@seen\endcsname{0.976 $\pm$ 0.002}
\expandafter\gdef\csname vfc@gemma270m-sip@unseen\endcsname{0.156 $\pm$ 0.018}
\expandafter\gdef\csname vfc@gemma270m-sip@out_of_scope\endcsname{0.969 $\pm$ 0.001}
\expandafter\gdef\csname vfc@gemma270m-sip@seeds\endcsname{3}

\newcommand{\vfcscored}[1]{%
  \ifcsname vfcscored@#1\endcsname\csname vfcscored@#1\endcsname\else\vfc@missing{scored}{#1}\fi}
\expandafter\gdef\csname vfcscored@out_of_scope\endcsname{533}
\expandafter\gdef\csname vfcscored@seen\endcsname{545}
\expandafter\gdef\csname vfcscored@unseen\endcsname{2{,}136}

\expandafter\gdef\csname vfclat@fg270m-ft@ttfc\endcsname{82\,ms}
\expandafter\gdef\csname vfclat@fg270m-ft@ttfcrange\endcsname{77--90\,ms}
\expandafter\gdef\csname vfclat@fg270m-ft@rss\endcsname{408\,MB}
\expandafter\gdef\csname vfclat@fg270m-ft@prompttokens\endcsname{16}
\expandafter\gdef\csname vfclat@fg270m-ft@prefill\endcsname{24\,ms}
\expandafter\gdef\csname vfclat@fg270m-ft@tokspersec\endcsname{116.1}
\expandafter\gdef\csname vfclat@fg270m-ft@reached\endcsname{19/20}
\expandafter\gdef\csname vfclat@fg270m-ft@runs\endcsname{3}
\expandafter\gdef\csname vfclat@fg270m-sip@ttfc\endcsname{2338\,ms}
\expandafter\gdef\csname vfclat@fg270m-sip@ttfcrange\endcsname{2290--2349\,ms}
\expandafter\gdef\csname vfclat@fg270m-sip@rss\endcsname{1208\,MB}
\expandafter\gdef\csname vfclat@fg270m-sip@prompttokens\endcsname{1921}
\expandafter\gdef\csname vfclat@fg270m-sip@prefill\endcsname{2148\,ms}
\expandafter\gdef\csname vfclat@fg270m-sip@tokspersec\endcsname{117.7}
\expandafter\gdef\csname vfclat@fg270m-sip@reached\endcsname{18/20}
\expandafter\gdef\csname vfclat@fg270m-sip@runs\endcsname{3}
\expandafter\gdef\csname vfclat@gemma270m-ft@ttfc\endcsname{87\,ms}
\expandafter\gdef\csname vfclat@gemma270m-ft@ttfcrange\endcsname{64--88\,ms}
\expandafter\gdef\csname vfclat@gemma270m-ft@rss\endcsname{407\,MB}
\expandafter\gdef\csname vfclat@gemma270m-ft@prompttokens\endcsname{16}
\expandafter\gdef\csname vfclat@gemma270m-ft@prefill\endcsname{25\,ms}
\expandafter\gdef\csname vfclat@gemma270m-ft@tokspersec\endcsname{118.6}
\expandafter\gdef\csname vfclat@gemma270m-ft@reached\endcsname{18/20}
\expandafter\gdef\csname vfclat@gemma270m-ft@runs\endcsname{3}
\expandafter\gdef\csname vfclat@gemma270m-sip@ttfc\endcsname{2277\,ms}
\expandafter\gdef\csname vfclat@gemma270m-sip@ttfcrange\endcsname{2248--2341\,ms}
\expandafter\gdef\csname vfclat@gemma270m-sip@rss\endcsname{1208\,MB}
\expandafter\gdef\csname vfclat@gemma270m-sip@prompttokens\endcsname{1921}
\expandafter\gdef\csname vfclat@gemma270m-sip@prefill\endcsname{2038\,ms}
\expandafter\gdef\csname vfclat@gemma270m-sip@tokspersec\endcsname{123.1}
\expandafter\gdef\csname vfclat@gemma270m-sip@reached\endcsname{19/20}
\expandafter\gdef\csname vfclat@gemma270m-sip@runs\endcsname{3}
\expandafter\gdef\csname vfclat@qwen06b-ft@ttfc\endcsname{141\,ms}
\expandafter\gdef\csname vfclat@qwen06b-ft@ttfcrange\endcsname{124--143\,ms}
\expandafter\gdef\csname vfclat@qwen06b-ft@rss\endcsname{1216\,MB}
\expandafter\gdef\csname vfclat@qwen06b-ft@prompttokens\endcsname{14}
\expandafter\gdef\csname vfclat@qwen06b-ft@prefill\endcsname{34\,ms}
\expandafter\gdef\csname vfclat@qwen06b-ft@tokspersec\endcsname{74.5}
\expandafter\gdef\csname vfclat@qwen06b-ft@reached\endcsname{20/20}
\expandafter\gdef\csname vfclat@qwen06b-ft@runs\endcsname{3}
\expandafter\gdef\csname vfclat@qwen06b-sip@ttfc\endcsname{5593\,ms}
\expandafter\gdef\csname vfclat@qwen06b-sip@ttfcrange\endcsname{5254--5666\,ms}
\expandafter\gdef\csname vfclat@qwen06b-sip@rss\endcsname{5126\,MB}
\expandafter\gdef\csname vfclat@qwen06b-sip@prompttokens\endcsname{1799}
\expandafter\gdef\csname vfclat@qwen06b-sip@prefill\endcsname{4502\,ms}
\expandafter\gdef\csname vfclat@qwen06b-sip@tokspersec\endcsname{25.0}
\expandafter\gdef\csname vfclat@qwen06b-sip@reached\endcsname{20/20}
\expandafter\gdef\csname vfclat@qwen06b-sip@runs\endcsname{3}
\expandafter\gdef\csname vfclat@qwen17b-ft@ttfc\endcsname{104\,ms}
\expandafter\gdef\csname vfclat@qwen17b-ft@ttfcrange\endcsname{91--360\,ms}
\expandafter\gdef\csname vfclat@qwen17b-ft@rss\endcsname{2683\,MB}
\expandafter\gdef\csname vfclat@qwen17b-ft@prompttokens\endcsname{14}
\expandafter\gdef\csname vfclat@qwen17b-ft@prefill\endcsname{95\,ms}
\expandafter\gdef\csname vfclat@qwen17b-ft@tokspersec\endcsname{23.8}
\expandafter\gdef\csname vfclat@qwen17b-ft@reached\endcsname{18/20}
\expandafter\gdef\csname vfclat@qwen17b-ft@runs\endcsname{3}
\expandafter\gdef\csname vfclat@qwen17b-sip@ttfc\endcsname{10206\,ms}
\expandafter\gdef\csname vfclat@qwen17b-sip@ttfcrange\endcsname{10039--10319\,ms}
\expandafter\gdef\csname vfclat@qwen17b-sip@rss\endcsname{6514\,MB}
\expandafter\gdef\csname vfclat@qwen17b-sip@prompttokens\endcsname{1799}
\expandafter\gdef\csname vfclat@qwen17b-sip@prefill\endcsname{8476\,ms}
\expandafter\gdef\csname vfclat@qwen17b-sip@tokspersec\endcsname{15.4}
\expandafter\gdef\csname vfclat@qwen17b-sip@reached\endcsname{20/20}
\expandafter\gdef\csname vfclat@qwen17b-sip@runs\endcsname{3}

\newcommand{\vfcratio}[2]{%
  \ifcsname vfcratio@#1@#2\endcsname\csname vfcratio@#1@#2\endcsname\else\vfc@missing{#1}{#2}\fi}
\expandafter\gdef\csname vfcratio@fg270m@ttfc\endcsname{29}
\expandafter\gdef\csname vfcratio@fg270m@rss\endcsname{3.0}
\expandafter\gdef\csname vfcratio@fg270m@extraprompt\endcsname{1{,}905}
\expandafter\gdef\csname vfcratio@gemma270m@ttfc\endcsname{26}
\expandafter\gdef\csname vfcratio@gemma270m@rss\endcsname{3.0}
\expandafter\gdef\csname vfcratio@gemma270m@extraprompt\endcsname{1{,}905}
\expandafter\gdef\csname vfcratio@qwen06b@ttfc\endcsname{40}
\expandafter\gdef\csname vfcratio@qwen06b@rss\endcsname{4.2}
\expandafter\gdef\csname vfcratio@qwen06b@extraprompt\endcsname{1{,}785}
\expandafter\gdef\csname vfcratio@qwen17b@ttfc\endcsname{98}
\expandafter\gdef\csname vfcratio@qwen17b@rss\endcsname{2.4}
\expandafter\gdef\csname vfcratio@qwen17b@extraprompt\endcsname{1{,}785}
\expandafter\gdef\csname vfcttfcratiorange\endcsname{26--98}
\expandafter\gdef\csname vfcrssratiorange\endcsname{2.4--4.2}
\expandafter\gdef\csname vfcratiopairs\endcsname{4}
\expandafter\gdef\csname vfcextrapromptrange\endcsname{1{,}785 to 1{,}905}

\makeatother

\input{tables/pvalues}
\input{tables/corpusmacros}

\title{From Fixed Keys to Readable Schemas: Small Language Models for Vehicle Agent Function Calls}

\author{
  Hamed Jafarzadeh Asl \\
  Huawei Noah's Ark Lab, Canada \\
  \texttt{hamed.jafarzadeh.asl@h-partners.com} \\
  \And
  Yuanhao Yu \\
  Huawei Noah's Ark Lab, Canada \\
  \texttt{yuanhao.yu@huawei.com}
  \And
  Vahid Partovi Nia \\
  Huawei Noah's Ark Lab and Polytechnique Montreal, Canada \\
  \texttt{vahid.partovinia@huawei.com}
  \vspace{-1em}
}

\DeclareMathOperator*{\argmax}{arg\,max}

\begin{document}

\maketitle

\begin{abstract}
In-vehicle assistants must translate natural-language requests into accurate vehicle function calls under strict memory and latency constraints, making small language models (SLMs) attractive for on-device deployment. For such models, a key design choice is how the available function surface is presented. Two approaches are to represent each function with a dedicated Functional Token (FT) or provide function schemas directly in the prompt. FTs enable compact inference but are restricted to functions learned during training, whereas Schema-in-Prompt (SIP) can generalize to unseen functions at the cost of longer prompts and higher inference overhead. We introduce a benchmark of 9,822 single-turn examples spanning 79 vehicle functions derived from Android Automotive, including held-out functions and requests requiring refusal. We compare both approaches under matched fine-tuning across four SLMs from 270M to 1.7B parameters. On functions seen during training, scaling provides limited benefit: the 270M model can match the 1.7B model, and the highest Seen accuracy in our main grid occurs at 0.6B. On held-out functions, FT achieves zero accuracy by construction, whereas SIP generalizes and improves substantially with scale. On out-of-scope requests, FT can invoke an unavailable function it was trained to emit, while SIP more reliably refuses based on the functions offered. This flexibility comes with a higher inference cost, chiefly the latency of processing the schema prompt. Our theoretical analysis formalizes why only SIP can predict functions withheld as training targets and why longer schema contexts increase inference cost. Overall, function-surface representation, rather than model scale alone, determines the capabilities and failure modes of SLM-based vehicle function calling.
\end{abstract}

\section{Introduction}
\label{sec:introduction}

\begin{figure}[t]
  \centering
    \fbox{\includegraphics[height=0.122\textheight]{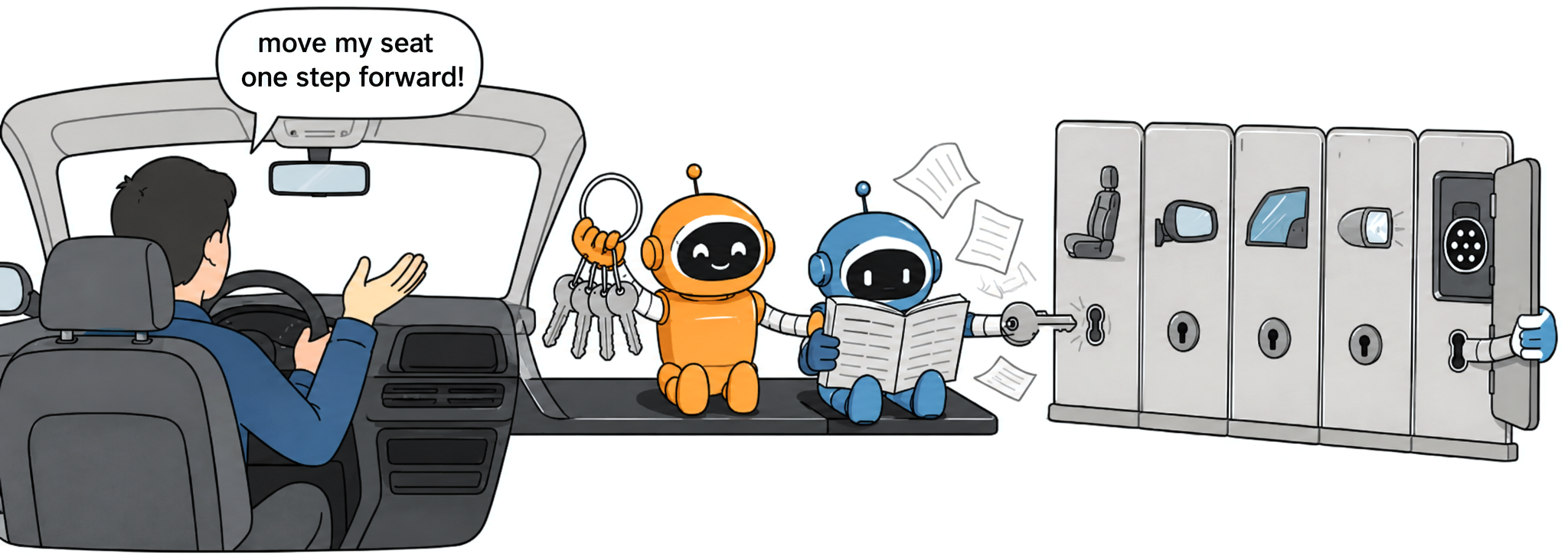}}
    \fbox{\includegraphics[height=0.122\textheight]{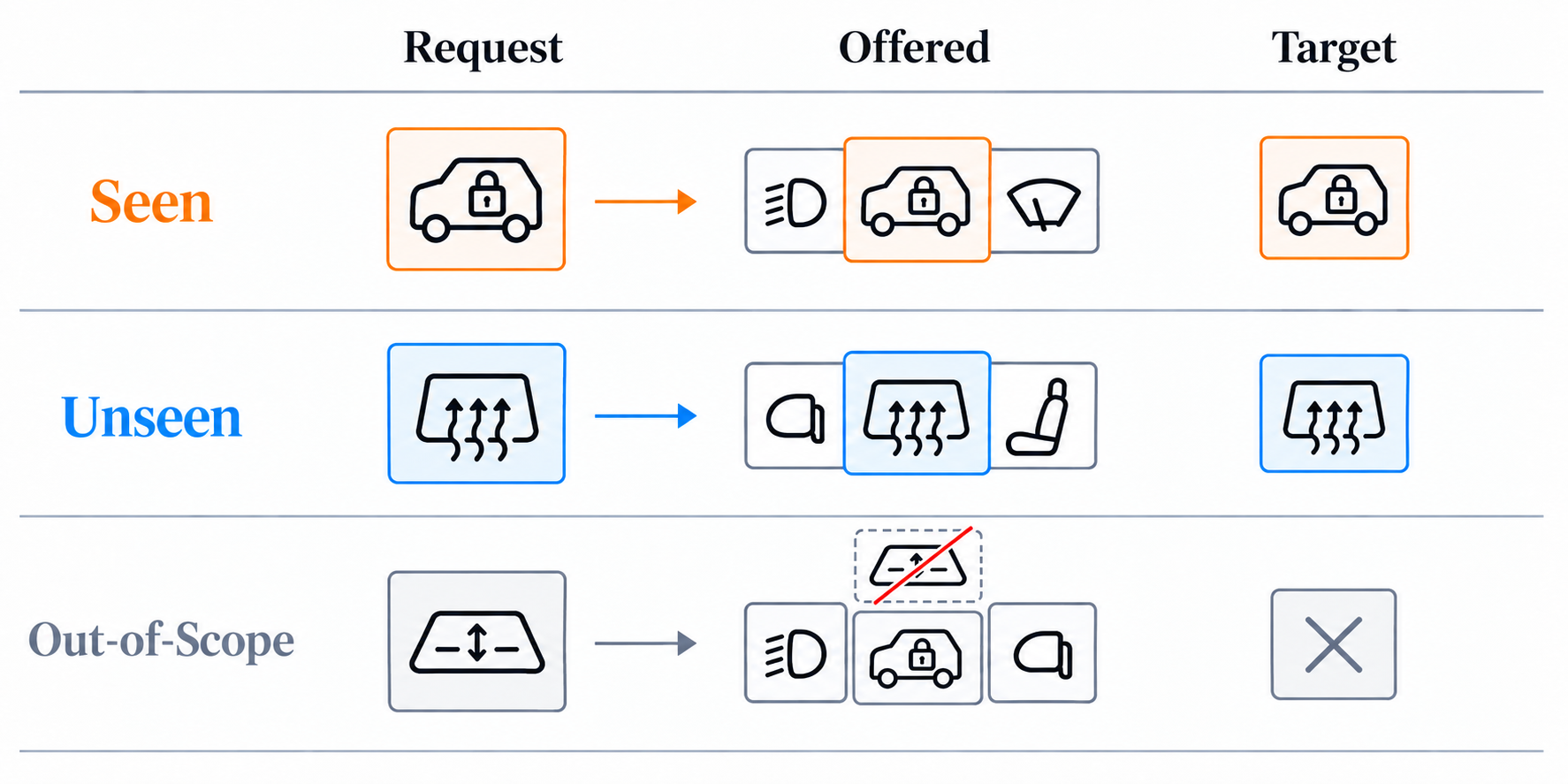}}
    \caption{
    Function-calling representations and evaluation splits.
    Left: Functional Token represents functions as fixed vocabulary entries, while Schema-in-Prompt conditions on offered function schemas at inference time.
    Right: \emph{Seen} functions appear as training targets, \emph{Unseen} functions are withheld as targets, and \emph{Out-of-Scope} requests have no suitable offered function and require refusal.
    }
    \vspace{-1.6em}
  \label{fig:examples}
\end{figure}

Natural-language control lets drivers operate a growing number of vehicle functions, many otherwise reachable only through touchscreen menus, without navigating those menus while driving. Reliable vehicle function calling requires an in-vehicle assistant to map a natural-language request to the correct vehicle action, including the appropriate function and arguments, or to refuse when the request cannot be satisfied. For example, ``move my seat one step forward'' should invoke the corresponding seat-control function with the correct arguments, as illustrated in Figure~\ref{fig:examples} (left). Unlike conventional text generation, an incorrect prediction may directly actuate physical hardware. Moreover, the available function surface can vary across vehicle models, model years, trim levels, and software updates, while deployment is constrained by on-board memory and latency and cannot assume network connectivity, which rules out relying on a cloud-hosted model for vehicle controls. These conditions make small language models (SLMs) attractive for on-device execution, but raise an important design question: \emph{how should the available functions be represented to a small model?}

We study two representations with fundamentally different properties. A \emph{Functional Token} (FT) approach~\citep{hao2023toolkengpt,octopusv2,baseline2025invehicle} assigns each supported function a dedicated vocabulary token learned during training. This representation is compact at inference time, but its task-level output space is tied to functions encoded during training, so supporting a new function requires adapting the representation and model. In contrast, \emph{Schema-in-Prompt} (SIP) supplies the available function schemas in the input context, as in the tool-definition interface of general-purpose function-calling models~\citep{xlam2024,hammer2024}; we use SIP as a descriptive label for this standard representation, not as a new method. The model interprets these schemas at inference time and selects among the functions offered for the current vehicle. SIP can therefore admit functions never used as training targets, at the cost of longer prompts and additional inference computation. The resulting question is not simply whether a larger SLM produces better function calls, but how the representation of the function surface affects generalization, refusal, and deployment cost.

Existing function-calling work explores several ways to adapt language models for tool use, including API selection, call construction, and invocation decisions. For compact deployment, Octopus v2 assigns functions dedicated vocabulary entries~\citep{octopusv2}, while other work investigates function masking, retrieval, low-rank adaptation, dynamic function selection, and action-specialized models~\citep{hammer2024,tinyagent2024,lessismore2024,xlam2024}. Both representations are complementary to structured-output and grammar-constrained decoding~\citep{willard2023efficient,geng2023grammar}, which guarantee well-formed calls and can be combined with either; a constraint built from the offered set can also block unavailable functions, but it cannot let an FT model predict a function for which it has no token. The closest prior work for in-vehicle function calling prunes a general language model and represents eight vehicle functions using dedicated vocabulary tokens~\citep{baseline2025invehicle}. Its proprietary function set and private dataset prevent a controlled numerical comparison, while the lack of a public vehicle-specific benchmark makes it difficult to evaluate how alternative representations behave as the supported function surface changes.
Language interfaces for vehicles have also been studied in adjacent settings. Talk2Car grounds passenger commands in street scenes~\citep{talk2car2019}, DriveLM studies visual question answering for driving~\citep{drivelm2023}, and Talk2Drive translates natural-language commands into driving controls~\citep{talk2drive2023}. These works primarily address perception, reasoning, planning, or driving control. We instead study single-turn selection and refusal over documented in-vehicle functions, focusing on how SLMs behave when the available function surface differs from the functions they were trained to execute.

We construct a vehicle function-calling benchmark from the vehicle-property API of Android Automotive OS (\texttt{VehiclePropertyIds}), which documents controllable properties such as seat position, climate control, and door locks, together with the cabin areas they apply to and their legal values. We convert these controllable properties into function schemas and separate functions observed as training targets, functions withheld as targets, and requests for which no suitable offered function exists. We use this benchmark to compare FT and SIP under matched training settings across SLMs from 270M to 1.7B parameters. Our results show that the two representations behave similarly on functions observed during training but diverge when the available function surface changes, revealing a trade-off between the flexibility of schema-conditioned function selection and its inference overhead.
Our contributions are:
\begin{enumerate}[itemsep=1pt, parsep=0pt, topsep=1pt]
\item[(i)] a reproducible vehicle function-calling benchmark derived from a documented Android Automotive function surface, with separate Seen, Unseen, and Out-of-Scope evaluation;

\item[(ii)] a controlled comparison of Functional Token and Schema-in-Prompt representations across SLMs from 270M to 1.7B parameters, including their function-selection, refusal, latency, and memory behavior;

\item[(iii)] a theoretical analysis explaining the difference in task-level output support between the two representations, how schema-mediated generalization to Unseen Functions becomes possible, and why this flexibility introduces additional inference cost.
\end{enumerate}

\section{Vehicle Function-Calling Benchmark}
\label{sec:benchmark}
\vspace{-0.3em}

We construct a benchmark for Single-Turn Function Calling from Android Automotive \texttt{VehiclePropertyIds}~\citep{vehiclepropertyids}. These specifications describe controllable vehicle properties together with the cabin areas (seats, windows, doors, mirrors) to which they apply and the values they accept. We convert the controllable properties into function schemas while preserving these constraints. For example, when a control can act on different seats or windows, the target area (such as the driver seat or a rear window) is represented as an argument. Documented numerical ranges and enumerated states define the legal argument values. Read-only properties that only report vehicle state are excluded. This produces a fixed, externally specified Function Surface rather than a collection of functions invented for the benchmark.

\paragraph{Generation and validation.}
For examples whose correct output is a Vehicle Function Call, we first select a function and sample legal argument values from its schema. The resulting call is validated programmatically before any natural-language request is generated. A language model (Llama-3.3-70B-Instruct) is then asked to express the validated call as a user request. The target function and arguments are therefore defined independently of the generated wording. We separately evaluate whether that wording is suitable for the benchmark. Generated requests are audited for \emph{Target Leakage}, where the wording directly exposes the target function identifier, screened by a second model from a different family (Mistral-Small-3.2-24B-Instruct), and manually inspected through a stratified \emph{Verified Slice}. The Verified Slice contains \vfccorpus{slicereviewed} examples across \vfccorpus{slicestrata} strata. Of these, \vfccorpus{slicefaithfulrate} are judged faithful to their targets and \vfccorpus{sliceunambiguousrate} are judged unambiguous. Appendix~\ref{app:benchmark_details} gives the complete construction and validation procedure. Requests are text: we assume an upstream speech recognizer and do not model recognition errors.

\paragraph{Evaluation design.}
Each benchmark example is associated with an \emph{offered set}, which defines the functions available for that example. \emph{Seen Functions} occur as targets during training. \emph{Unseen Functions} never occur as training targets. An Unseen Function is not undocumented or unvetted: it belongs to the vehicle's specified Function Surface and is offered through its validated schema. It models functionality that becomes available after the model was trained, for example through a software update or on another trim level, without retraining the model. The \vfccorpus{unseenfunctions} Unseen Functions form eight complete control families, chosen to span seat, window, mirror, and vehicle-wide controls and Boolean, stepped, bounded, unit-bearing numeric, and enumerated arguments; Appendix~\ref{app:offered_functions} lists them. Related functions are withheld together rather than holding out a single function while training on close counterparts from the same vehicle-control family. Their schemas may still appear as non-target candidates during training. SIP can therefore learn how to interpret offered schemas without being trained to select these held-out functions as answers.

\emph{Out-of-Scope Requests} require refusal because no suitable function is available. We construct two cases. Some requests ask for functionality absent from the complete Function Surface. Others request a valid function that is deliberately omitted from the offered set for that example. The second case tests whether the model recognizes that a function is unavailable rather than only identifying the operation requested by the user.

Each offered set contains \vfccorpus{offeredmin} to \vfccorpus{offeredmax} functions, with median \vfccorpus{offeredmedian}. Seen and Unseen schemas are mixed so that the composition of the offered set does not reveal the evaluation split or identify the answer by elimination. We also prevent equivalent requests from appearing on both sides of the training and evaluation partition. For globally unsupported functionality, different phrasings of the same underlying request type are assigned to the same partition.
Table~\ref{tab:corpus} summarizes the corpus with \vfccorpus{examples} examples, and the partition of its \vfccorpus{functions}-function surface into Seen, Unseen, and Out-of-Scope functions. The \vfccorpus{unseenfunctions} Unseen Functions are withheld as training targets, while the five Out-of-Scope functions are valid functions withheld from every offered set. Individual examples may additionally omit an otherwise available function from their own offered set.

\begin{table}[ht]
\centering
\begin{tabular}{lrrrr}
\toprule
Split & Examples & Training & Evaluation & Surface functions \\
\midrule
Seen & 5{,}521 & 4{,}975 & 546 & 56 \\
Unseen & 2{,}146 & 0 & 2{,}146 & 18 \\
Out-of-Scope & 2{,}155 & 1{,}622 & 533 & 5 \\
\midrule
Total & 9{,}822 & 6{,}597 & 3{,}225 & 79 \\
\bottomrule
\\
\end{tabular}

\caption{Benchmark composition and Function Surface partition. \emph{Surface functions} reports how the \vfccorpus{functions} functions are assigned to Seen, Unseen, and Out-of-Scope groups, not the number of distinct functions referenced by the examples in each row.}
\vspace{-1em}
\label{tab:corpus}
\end{table}

\section{Theoretical Analysis of Function-Surface Representation}
\label{sec:representation_analysis}

We theoretically characterize the distinction between Functional Token (FT) and Schema-in-Prompt (SIP) representations through their task-level output support. Let \(T\) denote the set of functions assigned dedicated Functional Tokens during training, and let \(\bot\) denote refusal. Under FT, the admissible task-level output set is fixed, denoted as $\mathcal{Y}_{\mathrm{FT}} = T \cup \{\bot\}.$
Therefore, a function \(f\notin T\) cannot be produced as an FT prediction unless the output representation is extended and the model is adapted to the new function.
SIP instead derives its admissible function set from the functions \(O_i\) offered for task instance \(i\), denoted as $
\mathcal{Y}_{\mathrm{SIP},i} = O_i \cup \{\bot\}.
$
A function withheld as a training target can therefore remain a valid prediction at inference time when its schema is included in \(O_i\). This difference also affects refusal behavior. SIP explicitly conditions on the functions available for the current task instance, whereas the evaluated FT formulation does not expose the row-specific offered set to the model.

The additional flexibility of SIP comes with a computational trade-off. Because SIP includes the serialized function schemas in the input context, its sequence length grows with the number and size of the offered schemas. For Transformer-based models, this increases prompt-processing cost and key--value cache requirements relative to FT, providing a theoretical basis for the latency and memory differences evaluated empirically in our device-proxy experiments. Appendix~\ref{app:theory} provides the full theoretical development, including the task-level support argument, schema-mediated generalization, refusal behavior, and inference-cost analysis.

\section{Experimental Setup}
\label{sec:setup}

\paragraph{Models and adaptation.}
We evaluate four compact open-weight models: Gemma~3-270M~\citep{gemma3technical2025}, FunctionGemma-270M~\citep{functiongemma2025}, Qwen3-0.6B, and Qwen3-1.7B~\citep{qwen3technical2025}. The two 270M models provide a same-scale comparison of general-purpose and function-calling-specialized pretraining, while the Qwen3 pair provides a within-family comparison of model capacity. Each model is adapted separately with Functional Token (FT) and Schema-in-Prompt (SIP) representations using the same training corpus. The 270M and 0.6B configurations use full fine-tuning, while Qwen3-1.7B uses LoRA~\citep{lora2021} in the main grid. Appendix~\ref{app:ablations} examines full fine-tuning for Qwen3-1.7B SIP and FunctionGemma's native function-calling prompt format.

\paragraph{Training and evaluation.}
Matched FT and SIP configurations use the same training examples and optimization recipe, with three independent seeds per configuration. We report exact Vehicle Function Call accuracy for Seen and Unseen Functions and correct-refusal accuracy for Out-of-Scope Requests. For Unseen Functions, we additionally report \emph{Execution}, the rate at which the model produces any valid Vehicle Function Call; the gap between Execution and exact accuracy therefore captures incorrect execution. For Out-of-Scope Requests, \emph{Over-trigger} measures the rate at which the model produces a Vehicle Function Call instead of refusing. Optimization, decoding, parsing, metric, and aggregation details are provided in Appendix~\ref{app:experimental_details}.

\paragraph{Device Proxy.}
Accuracy is evaluated with the fine-tuned checkpoints without quantization. Latency and memory are measured separately on a \emph{Device Proxy}: the same checkpoints are converted to GGUF with Q4\_K\_M weight quantization and served with llama.cpp~\citep{llamacpp} on four threads of a Xeon 8580 CPU, with a 4,096-token context window for both representations. Each configuration is measured three times on the same 20 evaluation requests. We do not evaluate accuracy under quantization, so these measurements characterize the relative inference cost of the two representations rather than the accuracy of quantized models or latency on vehicle hardware (Appendix~\ref{app:efficiency}).

\section{Results}
\label{sec:results}

\begin{figure}[t]
    \centering
    \begin{subfigure}[t]{0.49\linewidth}
        \centering
        \begin{tikzpicture}
        \begin{axis}[
            width=\linewidth,
            height=0.78\linewidth,
            ymin=80,
            ymax=100,
            ytick={80,85,90,95,100},
            ylabel={Accuracy (\%)},
            xlabel={Model},
            symbolic x coords={G-270M,FG-270M,Qwen-0.6B,Qwen-1.7B},
            xtick=data,
            x tick label style={font=\scriptsize,rotate=20,anchor=north east},
            y tick label style={font=\scriptsize},
            label style={font=\small},
            enlarge x limits=0.12,
            grid=major,
            grid style={gray!20},
            line width=1pt,
            mark size=2.2pt,
            legend style={
                at={(0.5,0.43)},
                anchor=center,
                legend columns=2,
                draw=none,
                font=\scriptsize
            },
        ]

        \addplot[color=blue!75!black,mark=*] coordinates {
            (G-270M,97.5)
            (FG-270M,97.1)
            (Qwen-0.6B,97.9)
            (Qwen-1.7B,95.5)
        };
        \addlegendentry{Functional Token}

        \addplot[color=red!75!black,mark=square*] coordinates {
            (G-270M,97.6)
            (FG-270M,97.2)
            (Qwen-0.6B,98.7)
            (Qwen-1.7B,98.1)
        };
        \addlegendentry{Schema-in-Prompt}

        \end{axis}
        \end{tikzpicture}
        \caption{Seen functions.}
        \label{fig:seen-panel}
    \end{subfigure}
    \hfill
    \begin{subfigure}[t]{0.49\linewidth}
        \centering
        \begin{tikzpicture}
        \begin{axis}[
            width=\linewidth,
            height=0.78\linewidth,
            ymin=80,
            ymax=100,
            ytick={80,85,90,95,100},
            ylabel={Accuracy (\%)},
            xlabel={Model},
            symbolic x coords={G-270M,FG-270M,Qwen-0.6B,Qwen-1.7B},
            xtick=data,
            x tick label style={font=\scriptsize,rotate=20,anchor=north east},
            y tick label style={font=\scriptsize},
            label style={font=\small},
            enlarge x limits=0.12,
            grid=major,
            grid style={gray!20},
            line width=1pt,
            mark size=2.2pt,
            legend style={
                at={(0.5,0.43)},
                anchor=center,
                legend columns=2,
                draw=none,
                font=\scriptsize
            },
        ]

        \addplot[color=blue!75!black,mark=*] coordinates {
            (G-270M,81.6)
            (FG-270M,81.4)
            (Qwen-0.6B,83.1)
            (Qwen-1.7B,82.6)
        };
        \addlegendentry{Functional Token}

        \addplot[color=red!75!black,mark=square*] coordinates {
            (G-270M,96.9)
            (FG-270M,97.3)
            (Qwen-0.6B,98.4)
            (Qwen-1.7B,95.7)
        };
        \addlegendentry{Schema-in-Prompt}

        \end{axis}
        \end{tikzpicture}
        \caption{Out-of-scope requests.}
        \label{fig:out-of-scope-panel}
    \end{subfigure}
    \caption{Seen Function and Out-of-Scope accuracy of the Functional Token (FT) and Schema-in-Prompt (SIP) approaches. FT and SIP perform similarly on Seen Functions (left), whereas SIP consistently achieves higher correct-refusal accuracy on Out-of-Scope Requests (right). FG denotes FunctionGemma-270M and G denotes Gemma~3-270M.}
    \label{fig:seen-out-of-scope-panels}
\end{figure}
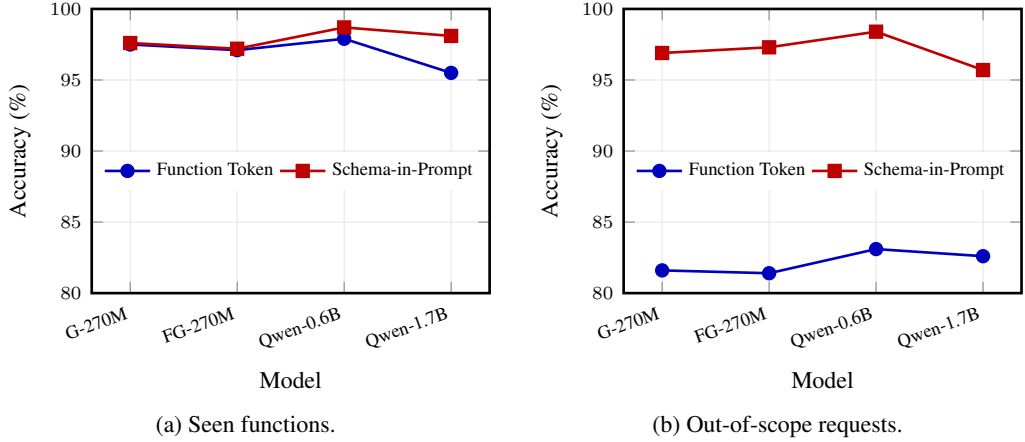

The experiments reveal a consistent division between representation and model capacity. Performance on functions observed during training is already high across the evaluated SLMs, while behavior under a changing Function Surface depends strongly on whether functions are encoded as learned outputs or supplied through schemas at inference time. Figures~\ref{fig:seen-out-of-scope-panels} and~\ref{fig:tradeoff} summarize these effects; the complete numerical grid is reported in Appendix~\ref{app:main_grid_results}.

\paragraph{Seen Function performance is already near saturation.}
Figure~\ref{fig:seen-out-of-scope-panels} (left) shows that both FT and SIP achieve high accuracy on Seen Functions across the evaluated models. Qwen3-0.6B obtains the highest mean Seen accuracy in the main grid, while the 270M configurations remain close to the larger models. For SIP, paired tests detect no significant difference between FunctionGemma-270M and Qwen3-1.7B on Seen Functions in any seed. This does not establish equivalence, but provides no evidence that the substantially larger configuration improves Seen Function performance in this comparison. Because the evaluated models span different families, we do not interpret the overall ordering as a general scaling law.

\paragraph{Without retraining, only SIP can predict Unseen Functions.}
FT's zero accuracy on Unseen Functions (Figure~\ref{fig:tradeoff}, left) follows from its output support (Section~\ref{sec:representation_analysis} and Appendix~\ref{app:task_level_output_support}) and serves as a reference point rather than a finding. Its practical consequence is that every function added after training, for example by a software update, requires extending the FT vocabulary and further training, and until then FT executes an incorrect function on some of these requests (Table~\ref{tab:failures}). SIP, by contrast, can predict an Unseen Function whenever its schema is offered, and achieves nonzero accuracy without retraining across all evaluated models. Within Qwen3, the 1.7B configuration substantially improves Unseen accuracy over 0.6B. At 270M, FunctionGemma also outperforms Gemma~3 on every seed, with a significant paired difference in each seed. Model capacity and prior function-calling specialization therefore improve schema interpretation, but only SIP provides the representational mechanism required to select a function that was never a training target.

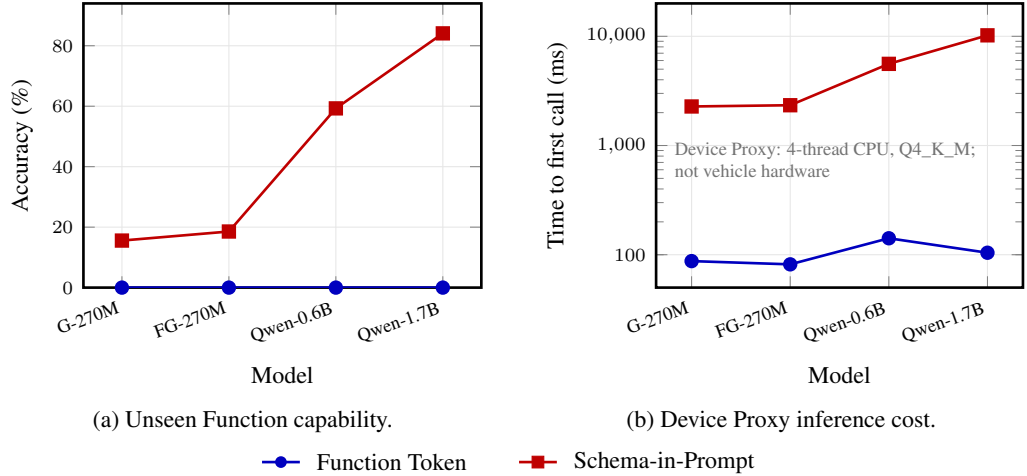
\begin{figure}[t]
\centering
%

\pgfplotsset{compat=1.18}

\begin{subfigure}[t]{0.49\linewidth}
    \centering
    \begin{tikzpicture}
    \begin{axis}[
        width=\linewidth,
        height=0.78\linewidth,
        ymin=0,
        ymax=94,
        ytick={0,20,40,60,80},
        ylabel={Accuracy (\%)},
        xlabel={Model},
        symbolic x coords={G-270M,FG-270M,Qwen-0.6B,Qwen-1.7B},
        xtick=data,
        x tick label style={
            font=\scriptsize,
            rotate=20,
            anchor=north east
        },
        y tick label style={font=\scriptsize},
        label style={font=\small},
        enlarge x limits=0.12,
        grid=major,
        grid style={gray!20},
        line width=1pt,
        mark size=2.2pt,
        clip mode=individual,
    ]

    \addplot[
        color=blue!75!black,
        mark=*
    ] coordinates {
        (G-270M,0)
        (FG-270M,0)
        (Qwen-0.6B,0)
        (Qwen-1.7B,0)
    };

    \addplot[
        color=red!75!black,
        mark=square*
    ] coordinates {
        (G-270M,15.559)
        (FG-270M,18.555)
        (Qwen-0.6B,59.285)
        (Qwen-1.7B,84.098)
    };

    \end{axis}
    \end{tikzpicture}
    \caption{Unseen Function capability.}
    \label{fig:unseen-panel}
\end{subfigure}
\hfill
\begin{subfigure}[t]{0.49\linewidth}
    \centering
    \begin{tikzpicture}
    \begin{axis}[
        width=0.94\linewidth,
        height=0.78\linewidth,
        ymin=50,
        ymax=20000,
        ymode=log,
        ytick={100,1000,10000},
        log ticks with fixed point,
        ylabel={Time to first call (ms)},
        xlabel={Model},
        symbolic x coords={G-270M,FG-270M,Qwen-0.6B,Qwen-1.7B},
        xtick=data,
        x tick label style={
            font=\scriptsize,
            rotate=20,
            anchor=north east
        },
        y tick label style={font=\scriptsize},
        label style={font=\small},
        enlarge x limits=0.12,
        grid=major,
        grid style={gray!20},
        line width=1pt,
        mark size=2.2pt,
    ]

    \addplot[
        color=blue!75!black,
        mark=*
    ] coordinates {
        (G-270M,87.419)
        (FG-270M,81.645)
        (Qwen-0.6B,141.492)
        (Qwen-1.7B,104.284)
    };

    \addplot[
        color=red!75!black,
        mark=square*
    ] coordinates {
        (G-270M,2277.303)
        (FG-270M,2337.991)
        (Qwen-0.6B,5592.686)
        (Qwen-1.7B,10205.538)
    };

    \node[
        anchor=west,
        font=\scriptsize,
        text=black!55,
        inner sep=1pt,
        align=left
    ] at (rel axis cs:0.04,0.45)
        {Device Proxy: 4-thread CPU, Q4\_K\_M;\\not vehicle hardware};

    \end{axis}
    \end{tikzpicture}
    \caption{Device Proxy inference cost.}
    \label{fig:ttfc-panel}
\end{subfigure}

\vspace{2pt}
\centerline{%
    \begin{tikzpicture}[baseline]

        \draw[
            color=blue!75!black,
            line width=1pt
        ] (0,0) -- (0.5,0);

        \fill[
            blue!75!black
        ] (0.25,0) circle (2.2pt);

        \node[
            anchor=west,
            font=\small
        ] at (0.6,0) {Functional Token};

        \draw[
            color=red!75!black,
            line width=1pt
        ] (3.4,0) -- (3.9,0);

        \node[
            fill=red!75!black,
            draw=red!75!black,
            inner sep=0pt,
            minimum size=4.4pt
        ] at (3.65,0) {};

        \node[
            anchor=west,
            font=\small
        ] at (4.0,0) {Schema-in-Prompt};

    \end{tikzpicture}%
}
\caption{Unseen Function capability and inference cost. Left, mean Unseen Function accuracy across three seeds. Right, median Device Proxy time to first call across three measurement runs. FG denotes FunctionGemma-270M and G denotes Gemma~3-270M. Device Proxy measurements are matched host measurements and do not represent latency on vehicle hardware.}
\label{fig:tradeoff}
\vspace{-1em}
\end{figure}

\paragraph{Execution exposes the consequence of unsupported targets.}
Unseen accuracy alone does not reveal whether an unsuccessful model refuses or attempts an incorrect action. Table~\ref{tab:failures} therefore reports \emph{Execution}, the rate of producing any valid Vehicle Function Call on an Unseen request, alongside exact accuracy. For FT, Unseen accuracy is zero, so any such execution is necessarily incorrect. For SIP, the gap between Execution and accuracy measures executions that do not exactly match the target call. This distinction is important for vehicle control: incorrect execution can actuate an unintended function or arguments, whereas refusal leaves the request unexecuted.

\paragraph{SIP better conditions refusal on the available Function Surface.}
Figure~\ref{fig:seen-out-of-scope-panels} (right) shows a consistent FT--SIP separation on Out-of-Scope Requests. The difference is concentrated in requests for valid vehicle functions that are absent from the offered set. When the unavailable function has a trained Functional Token, FT frequently produces a call despite the function being unavailable, whereas SIP usually refuses. When the unavailable function is not readily producible by either representation, both methods refuse reliably. This decomposition supports the representational explanation: SIP observes the offered Function Surface for the current example, while the evaluated FT formulation does not. Full results are reported in Appendix~\ref{app:failure_decomposition}.

\paragraph{Schema-mediated generalization introduces an inference cost.}
The additional capability of SIP comes from placing the offered schemas in the input context and therefore requires more prompt processing. Figure~\ref{fig:tradeoff} (right) shows the corresponding Device Proxy time-to-first-call difference. Across matched model pairs, SIP adds \vfcextrapromptrange{} prompt tokens and, when the schemas are encoded for every request, requires \vfcttfcratiorange$\times$ the Device Proxy time to first call of FT. Its per-request memory overhead is small: both representations use the same pre-allocated context window, which already accommodates the schema tokens. The higher peak resident memory of SIP in Appendix~\ref{app:efficiency} mainly reflects the runtime's host-memory prompt cache, which retained the states of all measured prompts; in a check of Qwen3-0.6B with this cache disabled, SIP's peak memory falls close to that of FT. These measurements agree with the sequence-length analysis in Appendix~\ref{app:theory}. Full measurements and implementation-specific effects are reported in Appendix~\ref{app:efficiency}.

Appendix~\ref{app:ablations} further tests Qwen3-1.7B adaptation method and FunctionGemma prompt format. Neither ablation changes the central result: FT provides the more compact representation, while SIP supports generalization and availability-aware refusal when the Function Surface changes.

\paragraph{Deployment implications.}
Three considerations qualify these costs. First, the Device Proxy re-encodes the schemas for every request, whereas a vehicle's offered set changes only with software updates: because our SIP prompt places the schemas before the request, their key--value states can be computed once and reused~\citep{gim2024prompt}, removing most of the prefill time at the cost of a cache that grows with the number of schemas. Without such caching, a smaller retrieved candidate set~\citep{tinyagent2024}, or an accelerator, the measured SIP times to first call (\vfclat{gemma270m-sip}{ttfc} to \vfclat{qwen17b-sip}{ttfc}) are too slow for interactive use. Second, neither representation should actuate hardware unchecked: calls can be validated against the schema and vehicle state, safety-relevant functions can require confirmation, and an FT decoder can mask the tokens of unavailable functions. Such a mask blocks calls to unavailable functions but cannot make FT predict a function without a token; whether the masked probability mass moves to refusal or to another function remains to be measured. Third, the representations can be combined: a compact FT model can serve a stable core of frequent functions, with SIP as a fallback for functions added after training or requests that FT refuses.

\begin{table}[t]
\centering
\setlength{\tabcolsep}{3pt}
\begin{tabular}{llccc}
\toprule
Representation & Model & \multicolumn{2}{c}{Unseen Functions} & Out-of-Scope \\
\cmidrule(lr){3-4}
 & & Accuracy & Execution & Over-trigger \\
\midrule
Functional Token & Qwen3-1.7B & 0.0 & 6.0 & 17.4 \\
 & Qwen3-0.6B & 0.0 & 0.4 & 16.9 \\
 & FunctionGemma-270M & 0.0 & 1.5 & 18.4 \\
 & Gemma~3-270M & 0.0 & 1.5 & 18.4 \\
\midrule
Schema-in-Prompt & Qwen3-1.7B & 84.1 & 87.9 & 4.3 \\
 & Qwen3-0.6B & 59.3 & 61.7 & 1.6 \\
 & FunctionGemma-270M & 18.6 & 21.6 & 2.7 \\
 & Gemma~3-270M & 15.6 & 18.2 & 3.1 \\
\bottomrule
\\
\end{tabular}

\caption{
Behavioral rates (\%) averaged over three seeds. For Unseen Functions, \emph{Accuracy} requires an exact match to the target Vehicle Function Call, whereas \emph{Execution} is the rate of producing an executable call to an available function. Their difference measures non-exact executions. For Out-of-Scope Requests, \emph{Over-trigger} is the rate of producing a Vehicle Function Call instead of refusing.}
\label{tab:failures}
\vspace{-1em}
\end{table}

\section{Conclusion}

Reliable in-vehicle function calling requires a small on-device model to map driver requests to the correct Vehicle Function Call, or refuse them when appropriate. This must be done under strict deployment constraints, even when the available Function Surface changes after training. We compare two ways to represent this surface: Functional Token (FT), which encodes functions as learned vocabulary entries, and Schema-in-Prompt (SIP), which provides function schemas at inference time. Both approaches perform strongly on Seen Functions, with limited benefit from additional model capacity in this regime. Their behavior diverges when the Function Surface changes. FT cannot correctly predict Unseen Functions outside its learned output space and may instead substitute an incorrect known function. SIP generalizes substantially better to Unseen Functions and more reliably handles Out-of-Scope Requests. Our theoretical analysis explains this difference through the task-level output support of the two representations.

This flexibility comes with a deployment cost. SIP requires longer prompts, which substantially raise Device Proxy latency when the schemas are encoded for every request, while its per-request memory overhead is small; FT remains more compact at inference time. Function Surface representation is therefore a deployment decision, not only a modeling choice. FT is better suited to stable Function Surfaces under tight inference constraints, whereas SIP is better suited to systems that must accommodate changing functions without modifying the learned output representation. Our benchmark is synthetic, English-only, text-only, and single-turn, and efficiency is measured on a quantized CPU Device Proxy rather than vehicle hardware. Future work should extend the evaluation to natural and speech-recognized driver requests, multilingual and multi-turn interaction, size-matched comparisons across model families, and measurements on vehicle hardware. The protocol itself (schemas derived from a documented API, a function-level holdout, and availability-dependent refusal) is not specific to vehicles and applies to other domains with a typed function surface, such as smart-home or mobile-OS actions.

\newpage

\bibliographystyle{apalike}
\bibliography{references}
\newpage 
\appendix

\section{Benchmark Construction and Validation}
\label{app:benchmark_details}

This appendix expands the benchmark construction described in Section~\ref{sec:benchmark}. We describe how documented Android Automotive controls are converted into function schemas, how structured calls and natural-language requests are generated, how the benchmark partitions are constructed, and how the resulting synthetic corpus is validated.

\subsection{Function Surface Construction}
\label{app:function_surface}

The Function Surface is derived from Android Automotive \texttt{VehiclePropertyIds}~\citep{vehiclepropertyids}. We retain properties that correspond to controllable vehicle functionality and exclude properties that only report vehicle state. Each retained control is converted into a callable schema using constraints provided by the documented interface.
When a control applies to several cabin areas, such as different seats or windows, the area is represented explicitly as an argument. Documented numerical ranges, units, Boolean states, and enumerated modes determine the admissible values of the remaining arguments. The resulting schemas therefore preserve the structure and constraints of the source interface while expressing them in a form suitable for Vehicle Function Calling.
The Function Surface is fixed before corpus generation, and the same definitions are used during both data construction and evaluation.
The Function Surface contains \vfccorpus{functions} functions partitioned into Seen, Unseen, and Out-of-Scope groups. Seen and Unseen Functions may serve as Vehicle Function Call targets, with \vfccorpus{unseenfunctions} Unseen Functions withheld as training targets. The Out-of-Scope group contains five valid functions that are deliberately withheld from every offered set, so requests for these functions require refusal. Out-of-Scope Requests are not limited to these five. An ordinary Seen or Unseen function may also be omitted from a single example's offered set, which makes it unavailable for that example alone. This second construction is what allows the refusal analysis in Appendix~\ref{app:failure_decomposition} to separate unavailable functions that the FT representation holds a token for from those it does not.

\subsection{Representative Examples}
\label{app:benchmark_examples}

Figure~\ref{fig:benchmark_examples} shows representative Seen, Unseen, and Out-of-Scope examples. Each row contains a natural-language request, the offered functions for that example, and the expected Vehicle Function Call or refusal.
The examples also illustrate why the offered set is part of the task definition. Nearly equivalent requests may require different outputs when the appropriate function is available in one offered set but absent from another.

\begin{figure}[t]
\centering
{\footnotesize
\begin{tabular}{@{}p{0.15\linewidth}p{0.81\linewidth}@{}}
\toprule
\textbf{Seen} & Request: \emph{``Lock the boot.''} \\
 & Label: \texttt{door\_set\_lock(area="DOOR\_REAR", on=true)} \\
 & Offered: \texttt{seat\_move\_lumbar\_vertical}, \texttt{mirror\_set\_heat}, \texttt{rear\_fog\_lights\_set}, \texttt{front\_fog\_lights\_set}, \texttt{seat\_set\_ventilation}, \texttt{seat\_move\_lumbar\_side}, \texttt{hvac\_set\_defroster}, \texttt{door\_set\_lock}, \texttt{hvac\_set\_max\_defrost}, \texttt{ev\_set\_charge\_limit}, \texttt{seat\_move\_cushion\_side\_support}, \texttt{seat\_move\_depth}, \texttt{steering\_wheel\_set\_heat}, \texttt{turn\_signal\_set} \\
 & \footnotesize The target function was a training target. \hfill\texttt{\footnotesize 0-003954} \\
\addlinespace[4pt]
\textbf{Unseen} & Request: \emph{``Turn on the rear demister.''} \\
 & Label: \texttt{hvac\_set\_defroster(area="WINDOW\_REAR\_WINDSHIELD", on=true)} \\
 & Offered: \texttt{seat\_move\_lumbar\_side}, \texttt{rear\_fog\_lights\_set}, \texttt{headlights\_set}, \texttt{front\_fog\_lights\_set}, \texttt{ev\_set\_charging}, \texttt{adas\_set\_blind\_spot\_warning}, \texttt{seat\_move\_headrest\_height}, \texttt{hvac\_set\_fan\_direction}, \texttt{cruise\_control\_set\_time\_gap}, \texttt{mirror\_move\_horizontal}, \texttt{cruise\_control\_set\_target\_speed}, \texttt{ev\_set\_charge\_current\_limit}, \texttt{ev\_set\_charge\_limit}, \texttt{seat\_move\_lumbar\_fore\_aft}, \texttt{hvac\_set\_defroster}, \texttt{cruise\_control\_set\_enabled}, \texttt{seat\_set\_easy\_access}, \texttt{window\_set\_child\_lock} \\
 & \footnotesize The schema is offered; the function was never a training target. \hfill\texttt{\footnotesize 0-011994} \\
\addlinespace[4pt]
\textbf{Out-of-Scope} & Request: \emph{``Lock the trunk.''} \\
 & Label: \textit{decline} \\
 & Offered: \texttt{set\_head\_up\_display}, \texttt{mirror\_move\_horizontal}, \texttt{ev\_set\_charging}, \texttt{cruise\_control\_set\_enabled}, \texttt{turn\_signal\_set}, \texttt{steering\_wheel\_set\_heat}, \texttt{mirror\_move\_vertical}, \texttt{seat\_move\_lumbar\_side}, \texttt{ev\_set\_charge\_current\_limit}, \texttt{hvac\_set\_power}, \texttt{hvac\_set\_max\_defrost}, \texttt{ev\_set\_charge\_port}, \texttt{seat\_move\_headrest\_height}, \texttt{seat\_move\_lumbar\_vertical}, \texttt{ev\_set\_charge\_limit}, \texttt{window\_set\_child\_lock}, \texttt{front\_fog\_lights\_set}, \texttt{hvac\_set\_defroster}, \texttt{hvac\_set\_fan\_speed}, \texttt{cruise\_control\_set\_target\_speed}, \texttt{seat\_move\_lumbar\_fore\_aft}, \texttt{adas\_set\_cross\_traffic\_monitoring} \\
 & \footnotesize The vehicle has this function, but it is absent from the offered list. Only the offered list makes the correct answer a refusal. Note the first row: the same function, requested almost the same way, is a correct call there and a correct refusal here. \hfill\texttt{\footnotesize 0-019056} \\
\bottomrule
\end{tabular}
}
\caption{Representative benchmark examples. Seen and Unseen examples require a Vehicle Function Call when the appropriate function is available. An Out-of-Scope example requires refusal when no suitable function is present in the offered set.}
\vspace{-1em}
\label{fig:benchmark_examples}
\end{figure}

\subsection{Label-First Generation}
\label{app:generation_pipeline}

Examples with a Vehicle Function Call as the correct output are constructed by generating the structured target before the natural-language request. We first select a function from the Function Surface and sample argument values that satisfy its schema. The resulting Vehicle Function Call is checked programmatically for a valid function identity, required arguments, argument types, permitted values, numerical ranges, and applicable vehicle area.
Only after this validation is the structured call provided to the corpus-writing model, which generates a natural-language request intended to express the same operation. We use Llama-3.3-70B-Instruct as the corpus writer. The target call therefore does not depend on a language model inferring a label from previously generated text.
This ordering separates structured validity from language quality. Programmatic validation establishes that the target itself is a legal Vehicle Function Call. It does not establish that the generated request describes that call faithfully. We assess these language-level properties separately in Appendix~\ref{app:quality_control}.
The generation prompt provides the intended operation, cabin area, and semantic interpretation of argument values. It also varies the requested speaking style to reduce repeated templated phrasing. Generated requests are normalized and deduplicated before partitioning. This prevents repeated or punctuation-equivalent requests from receiving disproportionate weight or appearing independently in training and evaluation.

\subsection{Unseen Functions and Offered Sets}
\label{app:offered_functions}

An Unseen Function is a function that never appears as the correct target of a training example. We withhold related functions together rather than holding out an isolated function while training on closely related functions from the same control family. This reduces the possibility that Unseen evaluation can be solved through direct matching to a nearly equivalent training target.

The \vfccorpus{unseenfunctions} Unseen Functions form eight complete control families. The families were chosen to span seat, window, mirror, and vehicle-wide controls and the Boolean, stepped, bounded, unit-bearing numeric, and enumerated argument types of the Function Surface, so that Unseen performance does not depend on one unusual family:
seat lumbar support (\texttt{seat\_move\_lumbar\_fore\_aft}, \texttt{seat\_move\_lumbar\_side}, \texttt{seat\_move\_lumbar\_vertical});
mirror position (\texttt{mirror\_move\_horizontal}, \texttt{mirror\_move\_vertical});
defrosting (\texttt{hvac\_set\_defroster}, \texttt{hvac\_set\_max\_defrost});
window child lock (\texttt{window\_set\_child\_lock});
fog lights (\texttt{front\_fog\_lights\_set}, \texttt{rear\_fog\_lights\_set});
EV charging (\texttt{ev\_set\_charging}, \texttt{ev\_set\_charge\_port}, \texttt{ev\_set\_charge\_limit}, \texttt{ev\_set\_charge\_current\_limit});
cruise control (\texttt{cruise\_control\_set\_enabled}, \texttt{cruise\_control\_set\_target\_speed}, \texttt{cruise\_control\_set\_time\_gap});
and steering-wheel heating (\texttt{steering\_wheel\_set\_heat}).
The five Out-of-Scope surface functions form two further families, seat memory (\texttt{seat\_memory\_select}, \texttt{seat\_memory\_store}) and vehicle modes (\texttt{set\_valet\_mode}, \texttt{set\_engine\_idle\_auto\_stop}, \texttt{set\_parking\_brake\_auto\_apply}).

The schemas of Unseen Functions are intentionally not removed from training entirely. They may appear as non-target candidates in the offered sets of ordinary training examples, but they are never the correct training answer. SIP can therefore learn the general task of comparing requests with supplied schemas without learning to select the held-out functions as target outputs.
Each example contains an offered set of \vfccorpus{offeredmin} to \vfccorpus{offeredmax} functions, with median \vfccorpus{offeredmedian}. Offered sets contain multiple Seen and Unseen schemas. This removes two potential shortcuts. If Unseen examples contained only unfamiliar schemas, the candidate composition itself could reveal the split. If the target were the only unfamiliar schema among otherwise Seen functions, it could instead be selected by elimination.

\subsection{Out-of-Scope Requests}
\label{app:oos_construction}

Out-of-Scope Requests require refusal because no suitable function is available for the current example. We construct two forms.
The first requests functionality that cannot be satisfied by any function in the complete Function Surface. The corpus contains \vfccorpus{vehiclecannot} examples of this type.
The second requests functionality represented by a real function, but that function is absent from the offered set for the current example. The corpus contains \vfccorpus{notoffered} examples of this type. This case is particularly important because the request alone does not determine the correct response. A similar request may require execution when the function is available and refusal when it is absent.
Both types use the same mixed offered-set construction as the rest of the benchmark. The presence of unfamiliar schemas therefore does not itself provide a cue that the correct output is refusal.

\subsection{Training and Evaluation Partition}
\label{app:data_partition}

The function holdout and the example partition serve different purposes. The function holdout determines which functions may occur as training targets. The example partition determines which individual requests belong to training or evaluation.
No example targeting an Unseen Function is assigned to training. Out-of-Scope examples occur in both partitions because refusal is a behavior that must be represented during training.
For ordinary examples, partitioning is based on normalized request text. Duplicate or punctuation-equivalent requests therefore remain on the same side of the split.
Globally unsupported requests require an additional safeguard. Multiple natural-language requests can be generated from the same underlying unsupported capability. If these were partitioned only by wording, one phrasing could appear during training and another during evaluation. We therefore assign all phrasings of the same underlying unsupported request type to the same partition. Evaluation consequently tests refusal on held-out request types rather than only on alternative phrasings of request types observed during training.

\subsection{Quality Control}
\label{app:quality_control}

The quality-control procedure separates properties that can be checked deterministically from properties that require semantic judgement.

\paragraph{Structured-call validation.}
Every Vehicle Function Call used as a target is checked against its schema before language generation. The validator verifies function identity, required arguments, argument types, permitted values, numerical ranges, and applicable vehicle areas.

\paragraph{Target Leakage.}
Synthetic data can become artificially easy when the generated request directly exposes the identifier of its target function. We refer to this as \emph{Target Leakage}. Requests are checked for literal function identifiers and distinctive combinations of identifier tokens. Ordinary argument values are not treated as leakage by themselves because a user may naturally need to specify a temperature, level, or mode. Requests that trigger the leakage criterion are regenerated or removed. The same leakage criterion is applied before evaluation scoring.

\paragraph{Independent semantic screening.}
We use Mistral-Small-3.2-24B-Instruct as an independent semantic screen. It belongs to a different model family from both the corpus writer and the SLMs evaluated in this paper. For examples with a Vehicle Function Call as the target, the screen checks whether the generated request is consistent with the structured call. For refusal examples, the relevant available functions are also considered when assessing whether refusal is appropriate.
We do not treat this model-based screen as ground truth. Comparison against manual review showed that some semantic error judgements were unreliable, particularly for functions whose movement direction is represented through signed argument values. These judgements are therefore retained as diagnostics rather than used automatically to remove examples. Automatic filtering is restricted to leakage cases supported by the independent programmatic leakage check.

\paragraph{Verified Slice.}
We manually review a stratified Verified Slice containing \vfccorpus{slicereviewed} examples across \vfccorpus{slicestrata} strata defined by evaluation split, function group, and argument structure. For each sampled example, the reviewer checks whether the request faithfully expresses its assigned Vehicle Function Call or refusal and whether another call could satisfy the same request equally well. The review finds \vfccorpus{slicefaithfulrate} of the sampled examples faithful and \vfccorpus{sliceunambiguousrate} unambiguous.
Of the \vfccorpus{slicereviewed} reviewed examples, \vfccorpus{sliceblind} belong to the initial review pass. An additional \vfccorpus{sliceafterrepair} examples were reviewed after targeted corpus updates, and \vfccorpus{slicerepaired} of those had themselves been modified. We therefore treat the Verified Slice as evidence about the sampled corpus rather than as a fully blinded estimate or a guarantee over every generated example.

\subsection{Evaluation Filtering}
\label{app:evaluation_filtering}

The counts in Table~\ref{tab:corpus} describe the generated training and evaluation partitions before the final leakage filter is applied. Examples flagged by the supported leakage criterion are excluded before scoring.
All reported experimental metrics therefore use \vfcscored{seen} Seen, \vfcscored{unseen} Unseen, and \vfcscored{out_of_scope} Out-of-Scope examples as their respective denominators.

\subsection{Scope and Limitations}
\label{app:benchmark_scope}

The benchmark evaluates English, Single-Turn Function Calling. It does not require dialogue history, changing vehicle state, or sequences of dependent Vehicle Function Calls. This scope isolates the effect of Function Surface representation from multi-turn planning and state tracking.
The utterances are synthetically generated rather than collected from drivers. Programmatic validation establishes the legality of the structured targets. Leakage auditing, independent semantic screening, and manual review address complementary sources of language-level error. These checks do not establish that the synthetic utterance distribution matches naturally occurring in-vehicle requests. Evaluation on naturally collected requests, additional languages, and multi-turn interactions remains future work.

\section{Theoretical Analysis of In-Vehicle Function Calling}
\label{app:theory}

\subsection{Task-Level Output Support}
\label{app:task_level_output_support}

We formalize the difference between Functional Tokens (FTs) and Schema-in-Prompt (SIP) in terms of their task-level output support. Let \(\Sigma\) denote the tokenizer vocabulary and let \(\Sigma^\ast\) denote the set of finite token sequences over \(\Sigma\). A \emph{Function Class} is a finite set \(\mathcal{F}=\{f_1,\ldots,f_m\}\). Each function \(f\in\mathcal{F}\) has a schema \(S(f)\in\Sigma^\ast\) containing its name, natural-language description, argument names, argument types, and value constraints.

For an evaluation row \(i\), let \(x_i\in\Sigma^\ast\) denote the user request and \(O_i\subseteq\mathcal{F}\) the set of functions offered for that vehicle. Let \(\mathcal{C}(x_i)\subseteq\mathcal{F}\) denote the set of functions that can semantically satisfy request \(x_i\). The target function-selection decision is \(y_i\in O_i\cup\{\bot\}\), where \(\bot\) denotes refusal or abstention. Under our task definition,
\[
y_i=\bot
\quad\Longleftrightarrow\quad
\mathcal{C}(x_i)\cap O_i=\varnothing.
\]
Thus, an \emph{Out-of-Scope Request} is one for which no function capable of satisfying the request is available in the offered set. This includes both requests unsupported by the vehicle function class and requests whose required function exists in \(\mathcal{F}\) but is absent from \(O_i\). This analysis isolates function identity selection; argument generation is omitted for clarity and does not change the output-support distinction considered below. A \emph{Seen Function} is a function that appears as a target during training, whereas an \emph{Unseen Function} is withheld as a training target but may still appear among the schemas offered at inference time.

Let \(T\subseteq\mathcal{F}\) denote the set of Seen Functions, each of which is assigned a dedicated FT during training, and let
\[
\tau:T\rightarrow\Sigma
\]
be an injective mapping from each function to its dedicated token. Although the underlying language model generates sequences in \(\Sigma^\ast\), the evaluated FT decoder maps valid FT outputs to function identities in \(T\). Therefore, at the task level, the admissible FT output set is
\[
\mathcal{Y}_{\mathrm{FT}}=T\cup\{\bot\}.
\]
Let \(p_{\mathrm{FT}}(y\mid x_i)\) denote the task-level output distribution induced by the language model together with this FT decoding rule. Then
\[
p_{\mathrm{FT}}(f\mid x_i)=0
\qquad
\forall f\in\mathcal{F}\setminus T.
\]
This is a statement about the support of the evaluated representation rather than about semantic similarity or the raw token-level language-model distribution. A request for an Unseen Function may be linguistically close to requests for Seen Functions, and the model could in principle generate ordinary tokens resembling its name. However, because no dedicated FT maps to \(f\notin T\), such a sequence is not decoded as a valid prediction of \(f\) under the evaluated FT representation. Consequently, exact Unseen Function accuracy is zero by construction. The informative measurement is therefore the observed failure mode: (i) refusal, (ii) prediction of an incorrect Seen Function, or (iii) an invalid or unparsable output.

Schema-in-Prompt instead constructs the admissible function set from the schemas supplied at inference time. Let
\[
\mathcal{S}(O_i)\in\Sigma^\ast
\]
denote the serialized representation of the schemas associated with the functions in \(O_i\). If
\[
O_i=\{f_{i,1},\ldots,f_{i,r_i}\},
\]
then, for the ordering used by the prompting procedure,
\[
\mathcal{S}(O_i)
=
S(f_{i,1})\Vert S(f_{i,2})\Vert\cdots\Vert S(f_{i,r_i}),
\]
where \(\Vert\) denotes concatenation together with the required schema separators and formatting tokens.

As with FT, we express the SIP decision rule at the task level, after mapping generated token sequences to valid function identities or refusal:
\[
\hat{y}_{\mathrm{SIP}}(x_i,O_i)
=
\argmax_{y\in O_i\cup\{\bot\}}
p_{\mathrm{SIP}}
\!\left(
y\mid x_i,\mathcal{S}(O_i)
\right),
\]
where \(p_{\mathrm{SIP}}\) denotes the task-level output distribution induced by the language model and the SIP parsing rule.

The admissible SIP function set is therefore row-specific. If an Unseen Function \(f\notin T\) is included in \(O_i\), it remains a valid task-level output even though it was never observed as a training target. This difference in output support explains the qualitative difference in Unseen Function behavior. Under the evaluated FT representation, a function outside \(T\) has no corresponding valid Functional Token and therefore cannot be predicted. Under SIP, the same function can enter the task-level output support through its schema and may be selected by interpreting that schema at inference time.

The same distinction affects refusal behavior. The correct decision depends on whether any function capable of satisfying the request belongs to the offered set:
\[
\mathcal{C}(x_i)\cap O_i=\varnothing
\quad\Longrightarrow\quad
y_i=\bot.
\]
At the model level, this can be viewed through a compatibility score \(s_\theta(x,f)\) between request \(x\) and function \(f\). For example, a selective decision rule may reject when
\[
\max_{f\in O_i}s_\theta(x_i,f)<\gamma,
\]
for some rejection threshold \(\gamma\). This connects the task to classical reject rules and modern selective classification \citep{chow1970reject,geifman2017selective}. In the evaluated FT formulation, \(O_i\) is not explicitly exposed to the model. Consequently, an FT model may emit a Seen Function that is compatible with the request but unavailable in the current row. SIP explicitly provides \(O_i\) through its serialized schemas, allowing both function selection and refusal to condition on the functions actually available.

This additional flexibility incurs a computational cost. Let \(|z|\) denote the token length of sequence \(z\), and define
\[
L_{{\mathrm{SIP}}, \ \! i}
=
|x_i|
+
|\mathcal{S}(O_i)|
+
c_{\mathrm{SIP}},
\]
where \(c_{\mathrm{SIP}}\) denotes fixed prompt and output-formatting overhead. Similarly, the FT input length is approximately
\[
L_{{\mathrm{FT}}, \ \! i}
=
|x_i|
+
c_{\mathrm{FT}},
\]
where \(c_{\mathrm{FT}}\) denotes the corresponding fixed overhead. Since
\[
|\mathcal{S}(O_i)|
\approx
\sum_{f\in O_i}|S(f)|
\]
grows with both the number and length of the offered schemas, \(L_{{\mathrm{SIP}}, \ \! i}\) can be substantially larger than \(L_{{\mathrm{FT}}, \ \! i}\).

For a standard Transformer with dense self-attention, attention-score computation during prompt prefill scales quadratically with input sequence length \citep{vaswani2017attention}. Thus, considering the attention component alone, SIP incurs approximately
\[
\mathcal{O}\!\left((L_{{\mathrm{SIP}}, \ \! i})^2\right)
\]
prefill computation, while key--value cache storage grows linearly with the cached sequence length. With key--value caching, the attention computation for each subsequent autoregressive decoding step is approximately linear in the existing context length. Therefore, the longer schema-augmented SIP prompts increase both prompt-processing work and cache requirements relative to FT. This scaling behavior is consistent with our proxy-device measurements: SIP extends task-level output support to Unseen Functions, but the additional schema tokens increase time to first call relative to FT fine-tuning.

\subsection{Generalization through Schema-in-Prompt Learning}
\label{app:SIP_generalization}

We next consider how additional training examples can improve Schema-in-Prompt (SIP) generalization. The notation in the preceding subsection describes individual task rows \((x_i,O_i,y_i)\). To reason about training statistically, let \((X,\mathsf{O},Y)\) denote the corresponding random task variables, whose realizations have the same domains:
\[
X\in\Sigma^\ast,
\qquad
\mathsf{O}\subseteq\mathcal{F},
\qquad
Y\in\mathsf{O}\cup\{\bot\}.
\]
Let \(P_{\mathrm{tr}}\) denote the training distribution and let
\[
D_n=\{(x_i,O_i,y_i)\}_{i=1}^{n}
\]
be a training set of \(n\) examples sampled from \(P_{\mathrm{tr}}\). For the task-level analysis, let
\[
\ell(\hat{y},y)
=
\mathbf{1}\{\hat{y}\neq y\}
\]
denote the zero--one loss, irrespective of the token-level objective used to train the language model. For a task-level predictor \(h_\theta\), define the training-distribution population risk and empirical risk as
\[
R_{\mathrm{tr}}(\theta)
=
\mathbb{E}_{(X,\mathsf{O},Y)\sim P_{\mathrm{tr}}}
\left[
\ell\!\left(h_\theta(X,\mathsf{O}),Y\right)
\right],
\]
and
\[
\widehat{R}_n(\theta)
=
\frac{1}{n}
\sum_{i=1}^{n}
\ell\!\left(h_\theta(x_i,O_i),y_i\right).
\]

Under standard i.i.d.\ sampling and uniform-convergence assumptions, generalization bounds take the schematic form
\[
R_{\mathrm{tr}}(\hat{\theta}_n)
\leq
\widehat{R}_n(\hat{\theta}_n)
+
\epsilon_n(\mathcal{H},\delta),
\]
where \(\hat{\theta}_n\) denotes the learned parameters, \(\mathcal{H}\) is the hypothesis class, \(\delta\in(0,1)\) is a confidence parameter, and \(\epsilon_n(\mathcal{H},\delta)\) is a complexity-dependent estimation term. For finite-capacity hypothesis classes, such terms commonly exhibit dependence of the form
\[
\epsilon_n(\mathcal{H},\delta)
=
\mathcal{O}\!\left(
\sqrt{
\frac{
\mathcal{C}(\mathcal{H})+\log(1/\delta)
}{n}
}
\right),
\]
for an appropriate capacity measure \(\mathcal{C}(\mathcal{H})\)
\citep[Ch.~6]{shalev2014understanding}. We use this expression only as a learning-theoretic motivation for training-time example scaling rather than as a model-specific generalization guarantee for the neural predictor used in our experiments. In particular, increasing \(n\) can reduce the estimation component under the assumptions above, but the bound neither guarantees monotonic empirical improvement nor, by itself, establishes generalization to functions withheld as training targets.

The Unseen Function setting introduced in Appendix~\ref{app:task_level_output_support} requires an additional form of transfer. Let \(P_{\mathrm{te}}\) denote the evaluation distribution. In the held-out-function evaluation, \(P_{\mathrm{te}}\) may contain rows whose target \(f_u\in\mathcal{F}\setminus T\) never appears as a training target under \(P_{\mathrm{tr}}\). Thus, the relevant SIP generalization mechanism cannot be explained solely as estimation of a fixed mapping from requests to training-time function identities. Instead, SIP can exploit a relation between request language and schema language that is shared across functions.

To make this intuition explicit, consider the compatibility score introduced in Appendix~\ref{app:task_level_output_support} under the following factorization:
\[
s_\theta(x,f)
=
\phi_\theta(x)^\top\psi_\theta(S(f)),
\]
where
\[
\phi_\theta:\Sigma^\ast\rightarrow\mathbb{R}^{d},
\qquad
\psi_\theta:\Sigma^\ast\rightarrow\mathbb{R}^{d}
\]
embed a request and a function schema, respectively. This factorization is an analytical abstraction of request--schema compatibility rather than a claim about the internal architecture of the autoregressive language model.

Generalizing the fixed threshold \(\gamma\) used illustratively in Appendix~\ref{app:task_level_output_support}, function selection with a possibly context-dependent reject threshold can be expressed as
\[
h_\theta(x,O)
=
\begin{cases}
\displaystyle
\argmax_{f\in O}s_\theta(x,f),
&
\displaystyle
\max_{f\in O}s_\theta(x,f)
\geq
\gamma_\theta(x,O),
\\[6pt]
\bot,
&
\displaystyle
\max_{f\in O}s_\theta(x,f)
<
\gamma_\theta(x,O),
\end{cases}
\]
where \(\gamma_\theta(x,O)\) represents a possibly context-dependent rejection threshold. As in Appendix~\ref{app:task_level_output_support}, this is a task-level abstraction of the model's function-selection and refusal behavior.

Under this view, additional training examples can improve SIP insofar as they improve request--schema relations that transfer beyond the specific function identities observed as training targets. Paraphrase diversity can reduce sensitivity to surface wording. Examples spanning different argument values or constraints can help distinguish the conditions under which otherwise similar functions should be selected or refused. Examples containing competing offered functions provide negative comparisons, requiring the model to distinguish schemas that may share vocabulary but correspond to different calls.

This provides a mechanism through which Unseen Function accuracy can improve even when the held-out function never appears as a training target. In particular, training on Seen Functions can improve parameters that encode correspondences between request language and schema descriptions. At evaluation time, the same learned relation can then be applied to the schema \(S(f_u)\) of an Unseen Function \(f_u\notin T\), provided that \(f_u\in O_i\). Such transfer is enabled by the schema-mediated representation, but it is not guaranteed merely by increasing the number of training examples: the added examples must contain information that improves relations relevant to the evaluation distribution.

FTs behave differently under training-time example scaling because their task-level output support remains fixed. Recall that
\[
\mathcal{Y}_{\mathrm{FT}}
=
T\cup\{\bot\}.
\]
For a Seen Function \(f\in T\), additional training examples can improve discrimination among the existing outputs. They may also improve refusal behavior when out-of-scope status can be inferred from request-only cues represented during training. For an Unseen Function \(f_u\notin T\), however, additional examples over the existing FT representation leave
\[
f_u\notin\mathcal{Y}_{\mathrm{FT}}
\qquad\text{and hence}\qquad
p_{\mathrm{FT}}(f_u\mid x)=0.
\]
Thus, additional examples cannot make \(f_u\) a valid task-level FT output without extending the FT representation and training the model to associate a new dedicated token with that function. Moreover, because \(O\) is not explicitly exposed in the evaluated FT formulation, additional examples alone cannot enable the FT model to condition its refusal decision on the row-specific absence of an otherwise compatible function.

SIP instead couples its admissible function set to the offered set \(O\). Consequently, a previously unseen function can become a valid task-level output simply by supplying its schema at evaluation time. Training-time example scaling can then improve two different aspects of SIP performance: estimation over patterns represented in the training distribution and transfer of learned request--schema relations to functions that were not observed as training targets. The latter distinguishes schema-mediated generalization from learning to classify more accurately among a fixed set of outputs.

\subsection{Risk--Cost Trade-offs of In-Context Demonstrations}
\label{app:risk_cost_tradeoff}

Although our experiments do not vary the number of in-context demonstrations, this subsection extends the preceding analysis to characterize the more general risk--cost trade-off that arises when additional examples are supplied through the inference context.

We now consider examples supplied at inference time rather than during parameter training. Let
\[
\mathbf{E}_k
=
(e_1,\ldots,e_k),
\qquad
e_j=(x'_j,O'_j,y'_j),
\]
denote an ordered sequence of \(k\) in-context demonstrations. Let
\[
\mathcal{D}(\mathbf{E}_k)\in\Sigma^\ast
\]
denote their complete serialization in the prompt, including the requests, offered schemas, target outputs, and required formatting. The sequence is ordered because demonstration selection and ordering can affect in-context predictions
\citep{liu2022goodicl,rubin2022retrieval,lu2022fantastically}. Using the notation of the preceding subsections, the SIP decision for evaluation row \(i\) becomes
\[
\hat{y}^{(k)}_i
=
\argmax_{y\in O_i\cup\{\bot\}}
p_{\mathrm{SIP}}
\!\left(
y
\mid
x_i,
\mathcal{S}(O_i),
\mathcal{D}(\mathbf{E}_k)
\right).
\]
The model parameters remain fixed; only the conditioning context changes. This is the inference-only adaptation commonly studied as in-context learning
\citep{brown2020language,xie2021explanation,akyurek2022learning,vonoswald2023transformers}. Prior work has shown that demonstrations can provide information about the input distribution, label space, output format, and other task cues
\citep{min2022rethinking}.

Within the compatibility abstraction introduced above, conditioning on \(\mathbf{E}_k\) induces a context-dependent score \(s_{\theta,\mathbf{E}_k}\) and rejection threshold \(\gamma_{\theta,\mathbf{E}_k}\). For a call-bearing row \(y_i\in O_i\), define the decision margin
\[
M^{\mathrm{call}}_{i,k}
=
s_{\theta,\mathbf{E}_k}(x_i,y_i)
-
\max
\left\{
\gamma_{\theta,\mathbf{E}_k}(x_i,O_i),
\;
\max_{f\in O_i\setminus\{y_i\}}
s_{\theta,\mathbf{E}_k}(x_i,f)
\right\},
\]
where the maximum over an empty set is defined as \(-\infty\). A positive margin means that the target function both outranks every competing offered function and exceeds the rejection threshold. For a refusal row \(y_i=\bot\), define
\[
M^{\mathrm{ref}}_{i,k}
=
\gamma_{\theta,\mathbf{E}_k}(x_i,O_i)
-
\max_{f\in O_i}
s_{\theta,\mathbf{E}_k}(x_i,f).
\]
A positive refusal margin means that every offered function falls below the rejection threshold. Under this abstraction, demonstrations can improve a particular prediction by increasing the corresponding decision margin. There is no requirement, however, that an arbitrary additional demonstration increases that margin.

This view is complementary to information-theoretic analyses of in-context learning. For example, \citet[Theorem~4.7]{jeon2024information} derive an upper bound on expected in-context log-loss containing irreducible, meta-estimation, and in-context estimation terms. Their result provides a theoretical account of how information supplied through the context can reduce estimation uncertainty under their assumptions. It does not imply that arbitrary demonstrations monotonically improve prediction, nor does it directly account for demonstration selection, ordering, distraction, or deployment latency. We therefore separate the predictive effect of demonstrations from their computational cost.

Using the evaluation distribution \(P_{\mathrm{te}}\) defined above, let the task-level evaluation risk of a predictor \(h\) be
\[
R_{\mathrm{te}}(h)
=
\mathbb{E}_{(X,\mathsf{O},Y)\sim P_{\mathrm{te}}}
\left[
\ell\!\left(h(X,\mathsf{O}),Y\right)
\right].
\]
Let \(h_0\) denote the SIP predictor without in-context demonstrations, and let \(h_{\mathbf{E}}\) denote the predictor conditioned on demonstration sequence \(\mathbf{E}\). For a demonstration budget \(k\), let \(\mathcal{A}_{k}\) denote the set of admissible ordered demonstration sequences containing at most \(k\) examples, including the empty sequence. Define the best achievable evaluation risk under this budget as
\[
R_{\mathrm{te}}^\star(k)
=
\inf_{\mathbf{E}\in\mathcal{A}_{k}}
R_{\mathrm{te}}(h_{\mathbf{E}}).
\]
For a finite admissible demonstration pool, the infimum can be replaced by a minimum. Because the empty sequence belongs to \(\mathcal{A}_{k}\),
\[
R_{\mathrm{te}}^\star(k)
\leq
R_{\mathrm{te}}(h_0).
\]

Define the attainable predictive gain at budget \(k\) as
\[
\Delta_{\mathrm{gain}}(k)
=
R_{\mathrm{te}}(h_0)-R_{\mathrm{te}}^\star(k)
\geq 0.
\]
For the particular demonstration sequence \(\mathbf{E}_k\) supplied to the model, define its context gap as
\[
\Delta_{\mathrm{gap}}(\mathbf{E}_k)
=
R_{\mathrm{te}}(h_{\mathbf{E}_k})-R_{\mathrm{te}}^\star(k)
\geq 0.
\]
The context gap measures the excess predictive risk of the selected context relative to the best admissible context under the same demonstration budget. It may reflect suboptimal demonstration selection or ordering, irrelevant or conflicting examples, and other context-induced effects. It should not be interpreted solely as a penalty arising from sequence length.

By construction,
\[
R_{\mathrm{te}}(h_{\mathbf{E}_k})
=
R_{\mathrm{te}}(h_0)
-
\Delta_{\mathrm{gain}}(k)
+
\Delta_{\mathrm{gap}}(\mathbf{E}_k).
\]
Therefore, the selected demonstrations improve predictive risk relative to the no-demonstration predictor exactly when
\[
R_{\mathrm{te}}(h_{\mathbf{E}_k})
<
R_{\mathrm{te}}(h_0)
\quad\Longleftrightarrow\quad
\Delta_{\mathrm{gain}}(k)
>
\Delta_{\mathrm{gap}}(\mathbf{E}_k).
\]

The use of at most \(k\) demonstrations gives a useful monotonicity property. Since
\[
\mathcal{A}_{k}
\subseteq
\mathcal{A}_{k+1},
\]
it follows that
\[
R_{\mathrm{te}}^\star(k+1)
\leq
R_{\mathrm{te}}^\star(k),
\]
or equivalently,
\[
\Delta_{\mathrm{gain}}(k+1)
\geq
\Delta_{\mathrm{gain}}(k).
\]
Thus, increasing the demonstration budget cannot worsen the \emph{best achievable} predictive risk, because every context admissible under budget \(k\) remains admissible under budget \(k+1\). This does not imply monotonic performance for the particular context actually used. The observed risk
\[
R_{\mathrm{te}}(h_{\mathbf{E}_k})
\]
may increase or decrease with \(k\) because the context gap can vary with demonstration selection, relevance, and ordering.

In-context demonstrations also incur a computational cost. Using the input-length notation of Appendix~\ref{app:task_level_output_support}, define
\[
C_{i,k}
=
|x_i|
+
|\mathcal{S}(O_i)|
+
|\mathcal{D}(\mathbf{E}_k)|
+
c_{\mathrm{SIP}}.
\]
For a standard Transformer with dense self-attention, attention-score computation during prompt prefill scales quadratically with the input sequence length,
\[
\mathcal{O}\!\left(C_{i,k}^{\,2}\right),
\]
while other components of Transformer computation have different scaling behavior
\citep{tay2022efficienttransformerssurvey}. This describes the sequence-length dependence of dense self-attention rather than an exact model of wall-clock latency.

To keep predictive quality and deployment cost distinct, let
\[
K_{\mathrm{te}}(\mathbf{E})
=
\mathbb{E}_{(X,\mathsf{O},Y)\sim P_{\mathrm{te}}}
\left[
\kappa(X,\mathsf{O},\mathbf{E})
\right]
\]
denote the expected inference cost associated with demonstration sequence \(\mathbf{E}\), where \(\kappa\) may denote a theoretical computation proxy or a measured device quantity such as time to first call. Define the additional inference cost relative to the no-demonstration setting as
\[
\Delta_{\mathrm{cost}}(\mathbf{E}_k)
=
K_{\mathrm{te}}(\mathbf{E}_k)-K_{\mathrm{te}}(\varnothing).
\]

If predictive risk and deployment cost are to be optimized jointly, they can be combined explicitly through
\[
J_{\mathrm{te}}(\mathbf{E})
=
R_{\mathrm{te}}(h_{\mathbf{E}})
+
\lambda K_{\mathrm{te}}(\mathbf{E}),
\qquad
\lambda\geq 0,
\]
where \(\lambda\) converts inference cost into the units of the deployment objective and specifies the application-dependent trade-off between predictive quality and computational cost. Substituting the definitions above gives
\[
J_{\mathrm{te}}(\mathbf{E}_k)-J_{\mathrm{te}}(\varnothing)
=
-\Delta_{\mathrm{gain}}(k)
+
\Delta_{\mathrm{gap}}(\mathbf{E}_k)
+
\lambda\Delta_{\mathrm{cost}}(\mathbf{E}_k).
\]
Hence the selected demonstrations improve the joint deployment objective exactly when
\[
\Delta_{\mathrm{gain}}(k)
>
\Delta_{\mathrm{gap}}(\mathbf{E}_k)
+
\lambda\Delta_{\mathrm{cost}}(\mathbf{E}_k).
\]

This decomposition separates three effects of in-context demonstrations: the predictive gain made attainable by access to demonstrations, the predictive gap associated with the particular demonstrations supplied to the model, and the additional deployment cost of the longer context. It also clarifies why the empirically useful number of demonstrations need not increase monotonically. A larger demonstration budget expands the set of contexts from which a useful prompt can be constructed, but a particular longer context may provide little additional predictive benefit, may introduce a larger context gap, and incurs additional prompt-processing cost. For an in-vehicle assistant operating under latency and compute constraints, the relevant operating point is therefore determined by the balance between predictive improvement and the cost of additional context.

\section{Experimental Details and Additional Results}
\label{app:experimental_details}

This appendix provides the implementation details, evaluation definitions, and additional analyses supporting Sections~\ref{sec:setup} and~\ref{sec:results}. We first describe the training and inference setup, then define the reported metrics and aggregation procedures. We subsequently provide the complete main-grid results, paired statistical tests, refusal decomposition, ablations, and Device Proxy measurements.

\subsection{Training and Inference Details}
\label{app:training_details}

All model--representation pairs use the same training examples and optimization recipe, with three independently trained seeds per configuration. Training uses three epochs, a learning rate of \(2\times10^{-5}\), batch size 16, a maximum sequence length of 4096 tokens, and a warmup ratio of 0.03. The 270M and Qwen3-0.6B models use full fine-tuning. Qwen3-1.7B uses LoRA for both FT and SIP in the main experimental grid; the corresponding fully fine-tuned SIP configuration is evaluated separately in Appendix~\ref{app:ablations}.

Inference uses greedy decoding with at most 128 generated tokens. FT and SIP predictions are parsed using matched criteria so that neither representation benefits from more permissive handling of formatting deviations.

\subsection{Evaluation Metrics and Aggregation}
\label{app:evaluation_metrics}

All model configurations are evaluated on the scored benchmark examples defined in Appendix~\ref{app:evaluation_filtering}. The same evaluation examples are used for all models and seeds within each split.

For Seen and Unseen Functions, \emph{accuracy} is the fraction of scored examples for which the parsed prediction exactly matches the target Vehicle Function Call, including its function and arguments, under the common evaluation criteria. For Out-of-Scope Requests, accuracy is the fraction for which the model correctly refuses rather than producing a Vehicle Function Call.

For Unseen Function requests, we additionally report \emph{Execution}, the fraction of scored examples for which the model produces a parsed Vehicle Function Call to a valid function. Exact accuracy is therefore a subset of Execution, and the difference between Execution and accuracy captures \emph{incorrect execution}: cases in which a Vehicle Function Call is produced but does not exactly match the target call. Other unsuccessful outcomes, such as refusal or invalid output, are not counted as Execution. For Out-of-Scope Requests, \emph{Over-trigger} is the fraction of scored examples for which the model produces a Vehicle Function Call when the correct behavior is refusal.

Unless otherwise stated, accuracy and behavioral rates are first computed separately for each of the three independently trained seeds and then averaged across seeds. Statistical tests instead operate directly on paired example-level correctness for each seed, as described in Appendix~\ref{app:significance}. Device Proxy measurements use the separate aggregation procedure described in Appendix~\ref{app:efficiency}.

\subsection{Complete Main-Grid Results}
\label{app:main_grid_results}

Table~\ref{tab:main} provides the complete numerical accuracy results for all model--representation pairs in the main experimental grid. These values underlie the Seen Function and Out-of-Scope results visualized in Figure~\ref{fig:seen-out-of-scope-panels}, as well as the Unseen Function accuracy shown in the left panel of Figure~\ref{fig:tradeoff}. The figures emphasize the main performance trends, while Table~\ref{tab:main} provides the exact mean accuracies used for comparison.

\begin{table}[ht]
\centering
\begin{tabular}{llccc}
\toprule
Representation & Model & Seen & Unseen & Out-of-Scope \\
\midrule
Functional Token & Qwen3-1.7B & 95.5 & 0.0 & 82.6 \\
 & Qwen3-0.6B & \textbf{97.9} & 0.0 & \textbf{83.1} \\
 & FunctionGemma-270M & 97.1 & 0.0 & 81.4 \\
 & Gemma~3-270M & 97.5 & 0.0 & 81.6 \\
\midrule
Schema-in-Prompt & Qwen3-1.7B & 98.1 & \textbf{84.1} & 95.7 \\
 & Qwen3-0.6B & \textbf{98.7} & 59.3 & \textbf{98.4} \\
 & FunctionGemma-270M & 97.2 & 18.6 & 97.3 \\
 & Gemma~3-270M & 97.6 & 15.6 & 96.9 \\
\bottomrule
\\
\end{tabular}

\caption{Accuracy (\%) averaged over three seeds for the complete main experimental grid. Out-of-Scope accuracy denotes correct refusal. Qwen3-1.7B uses LoRA for both representations, while the smaller models use full fine-tuning.}
\label{tab:main}
\end{table}

\subsection{Paired Statistical Tests}
\label{app:significance}

Because the compared configurations are evaluated on the same examples, differences in aggregate accuracy alone do not capture whether the models succeed and fail on the same requests. We therefore use two-sided exact McNemar tests on paired example-level correctness.
For each seed, the test considers only examples on which the two compared configurations differ in correctness. Its \(p\)-value quantifies the evidence against the null hypothesis that the two directions of disagreement are equally likely. We use a significance level of \(0.05\). A small \(p\)-value therefore indicates a systematic paired difference between the two configurations. A non-significant result means that such a difference is not detected and does not establish statistical equivalence. The tests are performed separately for the three training seeds.

Table~\ref{tab:significance} reports the comparisons used to support two claims in Section~\ref{sec:results}. The FunctionGemma-270M SIP versus Qwen3-1.7B SIP comparison tests whether the substantially larger configuration improves Seen Function performance. The FunctionGemma-270M SIP versus Gemma~3-270M SIP comparison instead holds model scale approximately fixed and tests whether prior function-calling specialization changes Seen or Unseen performance.

\begin{table}[ht]
    \centering
\begin{tabular}{@{}llccc@{}}
\toprule
Comparison (Schema-in-Prompt) & Split & Seed 0 & Seed 1 & Seed 2 \\
\midrule
FunctionGemma-270M vs Qwen3-1.7B & Seen & $0.607$ & $0.267$ & $0.238$ \\
FunctionGemma-270M vs Gemma~3-270M & Seen & $0.453$ & $1.000$ & $0.581$ \\
FunctionGemma-270M vs Gemma~3-270M & Unseen & $4.5 \times 10^{-4}$ & $5.6 \times 10^{-6}$ & $4.8 \times 10^{-6}$ \\
\bottomrule
\\
\end{tabular}

    \caption{Two-sided exact McNemar \(p\)-values for paired SIP comparisons. Each column corresponds to one independently trained seed, and each test uses identical evaluation examples for the two configurations. Values below \(0.05\) indicate a detected paired difference at the stated significance level.}
    \label{tab:significance}
\end{table}

For FunctionGemma-270M versus Qwen3-1.7B, no significant difference is detected on Seen Functions in any seed. This provides no evidence that the larger configuration improves Seen performance in this comparison, but it does not establish equivalence.

For FunctionGemma-270M versus Gemma~3-270M, no significant difference is detected on Seen Functions. On Unseen Functions, FunctionGemma has higher accuracy and the paired difference is significant in every seed. This provides evidence that prior function-calling specialization benefits Unseen Function interpretation for this same-scale model pair. We do not generalize this result beyond the evaluated pair.

\subsection{Out-of-Scope Refusal Decomposition}
\label{app:failure_decomposition}

Out-of-Scope Requests arise for two reasons. The requested functionality may be absent from the complete Function Surface, or the required function may exist in the Function Surface but be absent from the offered set for the current example. The latter case directly tests whether refusal depends on recognizing current function availability, so we analyze these examples separately.

We divide these requests according to whether the requested but unavailable function is a Seen Function and therefore has a trained Functional Token. The grouping is determined from the benchmark Function Surface partition and not from model predictions.
The \(n\) shown for each group in Table~\ref{tab:refusal_decomposition} is the number of unique scored evaluation requests in that group. It is not pooled across seeds. Every model and each of its three seeds are evaluated on the same \(n\) requests. The reported refusal rates are the mean of the three per-seed correct-refusal rates.

\begin{table}[ht]
    \centering
\begin{tabular}{@{}llcc@{}}
\toprule
Unavailable-function group & Model & FT refusal & SIP refusal \\
\midrule
FT has a token ($n=92$) & Gemma~3-270M & 2.9 & 85.5 \\
 & FunctionGemma-270M & 4.7 & 89.5 \\
 & Qwen3-0.6B & 4.7 & 93.5 \\
 & Qwen3-1.7B & 5.1 & 81.2 \\
\midrule
FT has no token ($n=164$) & Gemma~3-270M & 99.2 & 99.8 \\
 & FunctionGemma-270M & 98.8 & 99.0 \\
 & Qwen3-0.6B & 99.6 & 99.6 \\
 & Qwen3-1.7B & 97.6 & 97.8 \\
\bottomrule
\\
\end{tabular}

    \caption{Correct-refusal rates (\%) for Out-of-Scope Requests whose required function exists in the Function Surface but is absent from the offered set. \(n\) denotes the number of unique scored evaluation requests in each availability group. Rates are averaged over three seeds.}
    \label{tab:refusal_decomposition}
\end{table}

When FT has a trained token for the unavailable function, its correct-refusal rate is substantially lower than that of SIP across all evaluated models. In these examples, the requested function belongs to the learned FT output space, but the evaluated FT input does not expose the row-specific offered set. FT therefore cannot directly condition its decision on the function's absence.

When FT has no trained token for the unavailable function, both representations refuse much more reliably. The aggregate Out-of-Scope accuracy therefore combines two qualitatively different cases. The large FT--SIP refusal difference is concentrated in examples where FT can produce the requested function but cannot observe that it is unavailable.

\subsection{Adaptation and Prompt-Format Ablations}
\label{app:ablations}

Two ablations test experimental choices that could otherwise confound interpretation of the main grid.

\begin{table}[t]
    \centering
\begin{tabular}{@{}llccc@{}}
\toprule
Model & SIP configuration & Seen & Unseen & Out-of-Scope \\
\midrule
Qwen3-1.7B & LoRA $r{=}16$ (main grid) & 98.1 & 84.1 & 95.7 \\
 & Full fine-tune & 99.0 & 75.4 & 98.6 \\
\midrule
FunctionGemma-270M & Uniform JSON Schema (main grid) & 97.2 & 18.6 & 97.3 \\
 & Native tool-calling format & 97.4 & 24.9 & 94.8 \\
\bottomrule
\\
\end{tabular}

    \caption{SIP ablations for the Qwen3-1.7B adaptation method and FunctionGemma prompt format. Accuracy (\%) is averaged over three seeds. Each ablation is shown with its corresponding main-grid configuration.}
    \label{tab:ablations}
\end{table}

\paragraph{Qwen3-1.7B adaptation.}
The main Qwen3-1.7B configuration uses LoRA, whereas the smaller models use full fine-tuning. We therefore repeat Qwen3-1.7B SIP using full fine-tuning. Its Unseen Function accuracy decreases relative to the LoRA main-grid configuration but remains above Qwen3-0.6B SIP. The within-Qwen ordering reported in the main results therefore remains under either adaptation method. The experiment does not establish that LoRA is generally superior to full fine-tuning. Full fine-tuning also improves Out-of-Scope accuracy relative to the LoRA configuration, indicating that adaptation choice contributes to the 1.7B refusal behavior.

\paragraph{FunctionGemma prompt format.}
The main grid uses a common SIP representation across model families. FunctionGemma also provides a native function-calling format, so we evaluate that format while holding the corpus and evaluation examples fixed. The native format improves Unseen Function accuracy but reduces Out-of-Scope accuracy. This indicates that the uniform SIP format suppresses some of FunctionGemma's schema-reading capability while the native format introduces a refusal trade-off. The improvement does not remove the larger gap between the 270M and Qwen SIP configurations.

\subsection{Device Proxy Measurements}
\label{app:efficiency}

All latency and memory measurements are Device Proxy measurements. They are collected with llama.cpp's \texttt{llama-server} (build b10472) on a Xeon 8580 CPU using four threads under Q4\_K\_M quantization, with a 4,096-token context window for both representations. Each measurement run sends the same 20 evaluation requests, after two warm-up requests, to one server process. No vehicle hardware is used. The same measurement procedure is applied to FT and SIP so that the comparison primarily reflects their relative inference cost rather than absolute in-vehicle latency.

\begin{table}[ht]
    \centering
{\small
\begin{tabular}{@{}llrrr@{}}
\toprule
Representation & Model & Prompt & Time to first & Peak RSS \\
 & & tokens & call (ms) & (MB) \\
\midrule
Functional Token & Qwen3-1.7B & 14 & 104 (91--360) & 2683 (2682--2691) \\
 & Qwen3-0.6B & 14 & 141 (124--143) & 1216 (1215--1338) \\
 & FunctionGemma-270M & 16 & 82 (77--90) & 408 (407--411) \\
 & Gemma~3-270M & 16 & 87 (64--88) & 407 (405--407) \\
\midrule
Schema-in-Prompt & Qwen3-1.7B & 1{,}799 & 10206 (10039--10319) & 6514 (6512--6516) \\
 & Qwen3-0.6B & 1{,}799 & 5593 (5254--5666) & 5126 (5126--5126) \\
 & FunctionGemma-270M & 1{,}921 & 2338 (2290--2349) & 1208 (1208--1209) \\
 & Gemma~3-270M & 1{,}921 & 2277 (2248--2341) & 1208 (1208--1209) \\
\bottomrule
\\
\end{tabular}}

    \caption{Device Proxy inference measurements for the main experimental grid. Time to first call and peak resident memory are reported as the median across three measurement runs, with the observed minimum--maximum range in parentheses. All measurements use the same Xeon 8580 CPU, four threads, and Q4\_K\_M quantization. These values do not represent latency or memory use on vehicle hardware. Peak RSS includes the runtime's host-memory prompt cache, which retains the states of the measured prompts and overstates the per-request memory cost of SIP (see text).}
    \label{tab:efficiency}
\end{table}

\emph{Prompt tokens} is the number of input tokens processed before generation begins. \emph{Time to first call} is the elapsed Device Proxy time, including prompt processing, until the generated output first forms a parseable Vehicle Function Call or a refusal. \emph{Peak RSS} is the maximum resident set size observed during the measurement run. Time to first call and peak RSS are summarized by the median over the three runs, with the observed range shown in parentheses.

SIP processes substantially longer prompts because the offered schemas are serialized in its input, whereas FT uses a compact request-only representation. The increased prompt-processing work accounts for most of the SIP latency difference and is consistent with the sequence-length analysis in Section~\ref{sec:representation_analysis}.
The especially large relative time difference for Qwen3-1.7B should not be interpreted as a scaling law. Its FT configuration reaches a parseable call using fewer decoded tokens than Qwen3-0.6B FT, reducing the FT time that forms the denominator of the ratio.

Peak RSS should be read with care. Both representations use the same 4,096-token context window, whose key--value cache the runtime allocates in full when the context is created, so the schema tokens of a single SIP request fit within memory that FT also reserves. The server additionally keeps a host-memory prompt cache that retains the states of previously processed prompts, even though prefix reuse is disabled for every measured request. Over the measured prompts this cache grows with prompt length, and it accounts for most of the SIP--FT difference in Table~\ref{tab:efficiency}: for Qwen3, whose FP16 key--value state takes 112\,KiB per token (28 layers with eight 128-dimensional key--value heads), 20 prompts of \vfclat{qwen06b-sip}{prompttokens} tokens amount to about 3.9\,GB. Repeating the Qwen3-0.6B SIP measurement with the prompt cache disabled brings its peak RSS close to that of Qwen3-0.6B FT. The peak-RSS ratios in Table~\ref{tab:efficiency} therefore overstate the per-request memory cost of SIP.

\end{document}